\documentclass[preprint,12pt]{elsarticle}

\usepackage{lineno,hyperref}
\usepackage{natbib}
\usepackage{geometry}
\usepackage{fleqn}
\usepackage{graphicx}
\usepackage{newtxtext,newtxmath}
\usepackage{hyperref}
\usepackage{xurl}
\usepackage[utf8x]{inputenc}

\hypersetup{
    colorlinks,
    linkcolor={red!50!black},
    citecolor={blue!50!black},
    urlcolor={blue!80!black},
    pdfauthor={Lesego E. Moloko} 
}

\usepackage{booktabs}
\usepackage[justification=centering]{caption}

\usepackage{framed}  
\usepackage{nomencl} 
\makenomenclature

\usepackage{color,soul}
\usepackage{subcaption}
\DeclareMathOperator*{\argminA}{arg\,min}
\DeclareMathOperator*{\argmaxA}{arg\,max}
\usepackage{bm}
\usepackage{amsmath}

\usepackage[version=4]{mhchem}   
\usepackage[detect-all]{siunitx} 

\graphicspath{{./figs/}}
\journal{Annals of Nuclear Energy}

\usepackage{multicol}
\renewcommand*{\nompreamble}{\begin{multicols}{2}}
\renewcommand*{\nompostamble}{\end{multicols}}
\begin{document}

\begin{frontmatter}

\title{Prediction and Uncertainty Quantification of SAFARI-1 Axial Neutron Flux Profiles with Neural Networks}

\author[NESCA,NCSU]{Lesego E. Moloko\corref{correspondingauthor}}
\ead{lesego.moloko@necsa.co.za}
\author[NESCA,NCSU]{Pavel M. Bokov}
\author[NCSU]{Xu Wu}
\author[NCSU]{Kostadin N. Ivanov}

\cortext[correspondingauthor]{Corresponding author}

\address[NESCA]{The South African Nuclear Energy Corporation SOC Ltd (Necsa) \\
	Building-1900, P.O. Box 582, Pretoria 0001, South Africa \\}

\address[NCSU]{Department of Nuclear Engineering, North Carolina State University    \\
	Burlington Engineering Laboratories, 2500 Stinson Drive, Raleigh, NC 27695 \\}

\begin{abstract}

Artificial Neural Networks (ANNs) have been successfully used in various nuclear engineering applications, such as predicting reactor physics parameters within reasonable time and with a high level of accuracy. Despite this success, they cannot provide information about the model prediction uncertainties, making it difficult to assess ANN prediction credibility, especially in extrapolated domains. In this study, Deep Neural Networks (DNNs) are used to predict the assembly axial neutron flux profiles in the SAFARI-1 research reactor, with quantified uncertainties in the ANN predictions and extrapolation to cycles not used in the training process. The training dataset consists of copper-wire activation measurements, the axial measurement locations and the measured control bank positions obtained from the reactor’s historical cycles. Uncertainty Quantification of the regular DNN models' predictions is performed using Monte Carlo Dropout (MCD) and Bayesian Neural Networks solved by Variational Inference (BNN VI). The regular DNNs, DNNs solved with MCD and BNN VI results agree very well among each other as well as with the new measured dataset not used in the training process, thus indicating good prediction and generalization capability. The uncertainty bands produced by MCD and BNN VI agree very well, and in general, they can fully envelop the noisy measurement data points. The developed ANNs are useful in supporting the experimental measurements campaign and neutronics code Verification and Validation (V\&V).

\end{abstract}

\begin{keyword}
Uncertainty Quantification \sep Deep Neural Networks \sep Bayesian Neural Networks \sep Monte Carlo Dropout
\end{keyword}

\end{frontmatter}




\begin{table*}[!t]
\begin{framed}
  \section*{Nomenclature}
    \begin{multicols}{2}
     \subsection*{Symbols}
        \begin{description}
        \itemsep-0.1em\small
        \item[\rm $\bm{b}$] vector of learnable biases
        \item[\rm $\mathcal{D}$] training dataset
        \item[\rm $D_{KL}$] Kullback-Leibler divergence
        \item[\rm $\mathbb{E}$] mathematical expectation
        \item[\rm $\mathcal{F}$] variational free energy
        \item[\rm $\bm{h}^{(\ell)}$] vector of outputs from layer $\ell$
        \item[\rm $i$] sample index
        \item[\rm $k_{\ell}$] number of nodes in $\ell$-th layer
        \item[\rm $\ell$] layer index
        \item[\rm $L$] depth of a neural network
        \item[\rm $\mathcal{L}_{\text{dropout}}$] loss function in MCD 
        \item[\rm $m$] dimensionality of the output space
        \item[\rm $N$] number of samples 
        \item[\rm $n$] dimensionality of the input space
        \item[\rm $p$] dropout probability
        \item[\rm $R^2$] coefficient of determination
        \item[\rm $t$] forward pass index
        \item[\rm $T$] number of forward passes
        \item[$\bm{W}$] learnable weights
        \item[$\bm{x}$] input vector
        \item[$\bm{y}$] vector of target outputs
        \item[$\hat{\bm{y}}$] vector of predicted outputs
        \item[\rm $\bm{Z}_\ell$] dropout matrix before the $\ell$-th layer
        \item[\rm $\theta$] parameters controlling a distribution family
        \item[\rm $\lambda $] weight decay parameter
        \item[\rm $\mu$] mean
        \item[\rm $\sigma$] standard deviation
        \item[\rm $\phi(\cdot)$] activation function
        \end{description}
      \subsection*{Abbreviations}
      \begin{description}
        \itemsep-0.1em\small
        \item[\rm ANN]      Artificial Neural Network
        \item[\rm BNN]      Bayesian Neural Network
        \item[\rm BOC]      Beginning Of Cycle
        \item[\rm BWR]      Boiling Water Reactor
        \item[\rm CI]       Confidence Interval
        \item[\rm DNN]      Deep Neural Network
        \item[\rm FNN]      Feedforward Neural Network
        \item[\rm MCD]      Monte Carlo Dropout
        \item[\rm MAE]      Mean Absolute Error
        \item[\rm MCMC]     Markov Chain Monte Carlo
        \item[\rm MLE]      Maximum Likelihood Estimation
        \item[\rm MSE]      Mean Squared Error
        \item[\rm MTR]      Material Testing Reactor
        \item[\rm NN]       Neural Network
        \item[\rm PPF]      Power Peaking Factor
        \item[\rm ReLU]     Rectified Linear Unit 
        \item[\rm RMSE]     Root Mean Square Error
        \item[\rm SAFARI-1]	South African Fundamental Atomic Research Installation, generation 1
        \item[\rm UQ]       Uncertainty Quantification
        \item[\rm VI]       Variational Inference 
        \item[\rm V\&V]     Verification and Validation
      \end{description}
    \end{multicols}
  \vspace{-2mm}
  \end{framed}
\end{table*}


\section{Introduction}

The power and neutron flux distribution in the reactor core are important from its safety and operation point of view. The knowledge of the neutron flux distribution is essential in determining various reactor physical parameters such as its absolute power, fuel burnup, power peaking factor (PPF), peak-clad fuel temperature, safety margins, as well as in the monitoring of the reactor operation. Furthermore, the determination of axial and radial neutron flux distribution is also important for fuel management, code verification and validation (V\&V) studies \cite{Jaradat2016,Snoj2011} and experiments, as well as for various applications such as medical isotope production \cite{Haffner2019}. A good knowledge of the  flux distribution is, therefore, crucial for an efficient operation of the reactors without violation of safety limits.

The above-mentioned physical parameters are strongly influenced by the operating conditions of the reactor, such as fuel loading, control rods axial position, etc. Thus they are continuously subjected to changes for every core reload as well as during the normal operation of the reactor. Most of the research reactors \cite{international2019iaea} and power reactors in operation \cite{Dias2016a} are equipped with online/offline detectors or utilize foil/wire activation for neutron flux measurement and/or mapping, which is then used in the estimation of the reactor's physical parameters as well as for neutronics code V\&V. For instance, in the SAFARI-1 research reactor, the axial thermal neutron flux measurements are based on the activation of natural copper wires, placed between fuel plates, along the axial height of each assembly in the core \cite{moloko2016benchmark,international2015iaea}.

Simulation codes are usually used to calculate core operational parameters, including those mentioned above. In most cases, these codes have been successfully validated by means of the measured experimental data. Nonetheless, it is important to note that both the simulation codes and experimental measurements (especially uncontrolled experiments) are subjected to limitations, unknown factors, approximations, etc. In such cases, surrogate models, based on historical flux measurements, can be potentially employed as an extra source of information about the expected flux shapes (in addition to the measurements and neutronic simulations). Amongst many techniques for constructing surrogate models, the Artificial Neural Networks (ANNs) are gaining popularity, due to their efficiency and versatility.

An ANN -- also called feedforward neural network (FNN), multi-layer perception, or simply neural network (NN) -- is an information processing model, which is trained to perform  non-linear mapping between a so-called input space into an output space. ANN has been proven to be a powerful data-driven modelling tool and as such, has been increasingly applied in many fields, including pattern recognition, data mining, computer vision and natural language processing  \cite{Demuth2014,Suzuki2011}.

ANNs have been used in nuclear engineering for various applications for more than twenty years, often to predict reactor physics parameters within reasonable time and computational resources while yielding a high level of accuracy. For example, in \cite{Souza2006a} ANNs were used to improve the accuracy of the predicted PPFs in real time and thus reduce uncertainty margins placed upon the reactor safety limits. The NNs used the position of control rods and signals of ex-core detectors as input data; the detailed dataset was published in~\cite{Souza2006b}. It was shown that the radial basis function network performs better than the ANNs and that the PPF safety margin can be decreased by as much as \SI{5}{\percent}. 

In \cite{Terman2018}, ANNs were utilized to determine the position of the control rods and fuel burn-up based on the in-core self-powered neutron detector flux measurements. A sensitivity study, conducted with different ANN architectures, showed that the Bayesian regularization back-propagation algorithm has the best performance in predicting reactor parameters. The study further showed that the developed NNs has a satisfactory noise resistance response when trained by noisy datasets. ANNs have also been used as surrogate models for reactor optimization in order to improve the computational efficiency. To this end, ANN-based surrogate models for various utilization and safety parameters of the SAFARI-1 research reactor were constructed in \cite{Schlunz2015c} and subsequently used in \cite{Schlunz2016b,Schlunz2018} in multi-objective optimization studies.

Highly heterogeneous and high-dimensional large reactor cores (e.g. Boiling Water Reactors -- BWRs) usually lack high order of symmetry, thus making it difficult to estimate their core neutronics parameters. This challenge has been addressed in \cite{Rabie2020} by using NNs with two or more hidden layers, referred to as deep neural networks (DNNs) to estimate PPFs, control rod bank levels, and the cycle length of the Ringhals-1 BWR unit. The DNN design process was assisted by employing a so-called hyperparameter optimization technique. Results of the study demonstrate a promising performance with a high level of accuracy in predicted PPF, control rod bank level and cycle length and thus prove the ability of the NN in estimating core parameters in large reactor cores, which usually lacks symmetry. In \cite{koo2019}, DNN was used to predict the reactor vessel water level during a postulated Loss-Of-Coolant Accident. The developed DNN model was found to be capable of predicting the reactor vessel water level and was considered to perform well enough to provide monitoring support information to operators during severe accidents.

Performing robust uncertainty quantification (UQ) of the approximation uncertainties in NN models can improve the credibility of NN applications in nuclear engineering. This is especially necessary when NN models are used to make generalizations, i.e.~when providing predictions in extrapolated domains beyond the training domain. Examples of UQ in an existing DNN model, including theoretical frameworks and applications, are discussed in \cite{gal2016dropout,blundell2015weight}. The Dropout technique \cite{Srivastava2014dropout}, used in many DNN models to avoid over-fitting, is employed as an approximate Bayesian Inference in \cite{gal2016dropout} and ``Bayes by Backprop'' (i.e. uncertainty on the weights of an NN) in \cite{blundell2015weight} to represent model uncertainty. Both studies show that UQ is indispensable for both classification and regression problems.

In \cite{moloko2021estimation}, the capability of measurement-based ANNs to predict the assembly axial neutron flux profiles, given a control rod bank position, was explored. Training data consisted of copper-wire activation measurements and the measured control bank positions obtained from the SAFARI-1 research reactor's historical cycles. Shallow ANNs were constructed using MATLAB's Neural Network Toolbox. The study adopted two widely used algorithms, that is, Levenberg-Marquardt and Bayesian Regularization algorithms for training NNs. The ANN performance was enhanced by employing data normalization and smoothing techniques. The ANNs yielded a reasonable level of accuracy when compared to measurements and applied to unseen cycle data. It was also demonstrated that the Bayesian Regularization algorithm outperforms the Levenberg-Marquardt algorithm in terms of robustness, generalization and prediction accuracy. In addition, a possible application of the ANNs for automated identification and correction of defective measurements was discussed.

In the subsequent study \cite{moloko2022UQ}, a larger dataset was used with a DNN model and the prediction/approximation uncertainties were quantified using Monte Carlo Dropout (MCD) techniques, as well as Bayesian Neural Networks (BNNs), solved by variational inference. The predicted assembly axial neutron flux profile (with uncertainty bands) results showed that the three different NN models (regular DNN, DNN solved with MCD and BNN) produce results that agree reasonably well among each other and with the measurement data, on cycles that are not used in the training process. Besides the excellent generalization capability, the uncertainty bands produced by MCD and BNN agree well, and in general, they can fully envelop the noisy measurement data points.

In both \cite{moloko2021estimation} and \cite{moloko2022UQ}, it was however observed that the performance of the trained NNs is affected by various parameters and choices made in the process of training. These include: a trade-off between the amount and the quality of training data; approaches selected for pre-processing of the training data and lack thereof; the design (hereafter referred to as architecture) of an NN which involves the choice of input and output parameters, activation functions, the number of hidden layers and neurons in each layer and, occasionally, the objective function; choice of an NN training tool and specific training and UQ approaches within the tool as well as various parameters and options of a particular training algorithm. Finally, miscellaneous limitations of a practical nature also may play an important role, such as availability of computational resources and scarcity and deficiencies of experimental data used in NN training.

It is therefore paramount for practical applications of the developed NN metamodel to ensure that, despite all these factors, the constructed NNs provide an accurate and consistent flux shape prediction. This task gets even more complicated, due to the above-mentioned deficiencies in measured data, especially a relatively large noise-to-signal ratio. This work can be regarded in this context as a cross-verification of different NN training and UQ approaches.   

The rest of this paper is organized as follows. In Section~\ref{sec:problem_definition}, we describe the copper-wire measurement and data pre-processing procedures, and discuss the sensitivity of measured flux profiles to control bank positions for the available historical data.  In Section~\ref{sec:theory}, we provide a brief overview of NN training and UQ theory as it pertains to this study, while Section \ref{sec:methodology} provides details on design and training of NNs. Results are presented and analysed in Section \ref{sec:results-discussion}. Section \ref{sec:conclusions} contains concluding remarks and recommendations for future studies.    

\section{Problem definition}
\label{sec:problem_definition}

\subsection{The SAFARI-1 experimental data}
\label{sec:experimentaldata}

\begin{figure}[!tb]
	\centering
	\includegraphics[width=.8\linewidth]{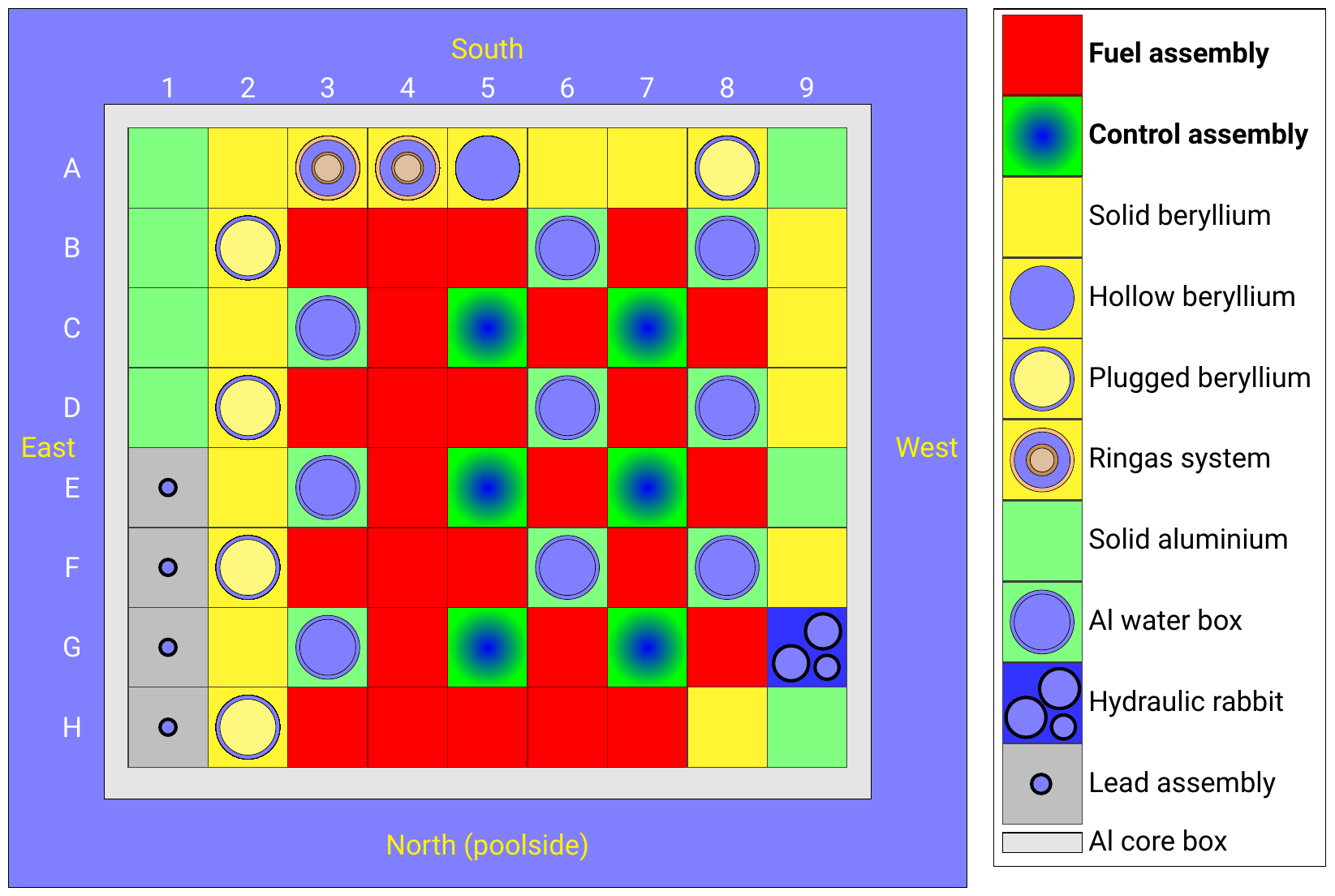}
	\caption{The SAFARI-1 core configuration \cite{moloko2021estimation}.}
	\label{figure:safaricore}
\end{figure}

SAFARI-1 (South Africa Fundamental Atomic Research Installation generation 1) is a 20 MW tank-in-pool-type material testing reactor (MTR) at Pelindaba, South Africa. The reactor is cooled and moderated by light water. Its $9\times8$ core lattice houses 26 MTR plate-type fuel assemblies, 6 follower-type control assemblies, a number of solid lead shield elements, solid and hollow aluminium filler elements, as well as several isotope production facilities. The core is reflected by beryllium elements on three sides with one side directly facing the reactor pool (poolside). A follower-type control assembly consists of a top absorber and bottom fuel-follower section, connected by an aluminium coupling piece. The fuel- and follower assemblies consist of 19 and 15 fuel plates respectively. The reactor is fuelled with \SI{19.75}{\percent} enriched Uranium Silicide (\ce{U3Si2Al}) fuel \cite{moloko2016benchmark, moloko2021estimation}. The SAFARI-1 core layout and the placement of fuel- and control assemblies are shown in Figure~\ref{figure:safaricore}.

The reactor is typically operated between 21 to 30 days with a 5-day shutdown and one extended 12-day shutdown per annum. Before the reactor start-up or beginning of each cycle, several experimental measurements are performed, in order to ensure that fuel management and thus safety reactor operation measures are adhered to. Of interest to this study is the measurement of the SAFARI-1 core thermal neutron flux distribution conducted by the irradiation/activation of natural copper wires. The experiment is hereinafter referred to as the copper-wire measurement.

The activation and measurement procedure entails preparation, loading, irradiation and retrieval of the copper wires. The sword-like, hollow aluminium blade containing a copper wire, is inserted along the active height of 32 assemblies, of which 26 are fuel assemblies and 6 are fuel-followers of the control assemblies. It is irradiated at a low reactor power of about \SI{200}{\kilo\watt} for controlled times \cite{moloko2021estimation}. For each of these measurements, the axial position of the control bank (6 control rods) is recorded.

Natural copper consists of two isotopes, that is \ce{^{63}Cu} and \ce{^{65}Cu} with abundances of \SI{69.15}{\percent} and \SI{30.85}{\percent}, respectively. It is activated according to reactions \ce{^{63}Cu(n,\gamma)^{64}Cu} and \ce{^{65}Cu(n,\gamma)^{66}Cu}. The \ce{(n,\gamma)} thermal reaction cross section of \ce{^{63}Cu} is approximately double that of \ce{^{65}Cu} and its product nuclide \ce{^{64}Cu} decays with a half-life of \SI{12.7}{\hour}. The product nuclide of \ce{^{65}Cu} has a half-life of only \SI{5.1}{\minute} and it decays away in just an hour, after which the activity is predominately attributed to the decay of \ce{^{64}Cu}. The wires are then removed and the radioactive \ce{\beta}-decay of the \ce{^{64}Cu} isotope with a half-life of \SI{12.7}{\hour}, is measured at 180 equidistant points along the length of the wires.

The measurement procedure takes approximately 5 minutes per wire and up to a total time of between 1 to 2 hours. The activity (or detector counts) for each of the 32 assemblies are saved in a data file for further processing and calculations of parameters of interest. This data is routinely used for the V\&V of reactor simulation tools, for the estimation of the safety-related parameters such as peak-clad temperatures in fuel, core power and neutron flux distribution, as well as to detect a potential fuel element misloading. Historical copper-wire measurements for each cycle and spanning years of reactor operations are available for analysis \cite{international2022iaea}.

\subsection{Sensitivity of the neutron flux profile to bank movement}
\label{subsec:flux_banks_sensitivity}

The practical applicability of the proposed NNs depends on whether or not the flux shape is a sufficiently sensitive function of the control bank position \cite{Garis1998}, i.e. there is actually a dependence to fit. The analysis herein seeks to illustrate and quantify the effect of the control bank position on the measured assembly axial neutron flux profiles, and thus deduce if this change in the bank position can be seen by the NNs models. In practice, the degree to which the flux variation can be associated with control placement is determined (limited) by the following factors:

\begin{enumerate}
  \item The range, in which the measured control bank axial position varied in considered historical cycles; this affects the magnitude of expected flux shape distortions;
  \item The difference in sensitivity of the flux profile in individual positions of the reactor core;
  \item The contribution of noise to the total variability of measured flux (signal-to-noise ratio).
\end{enumerate}
%

\begin{figure}[!tb]
  \centering
  \includegraphics[width=0.7\linewidth]{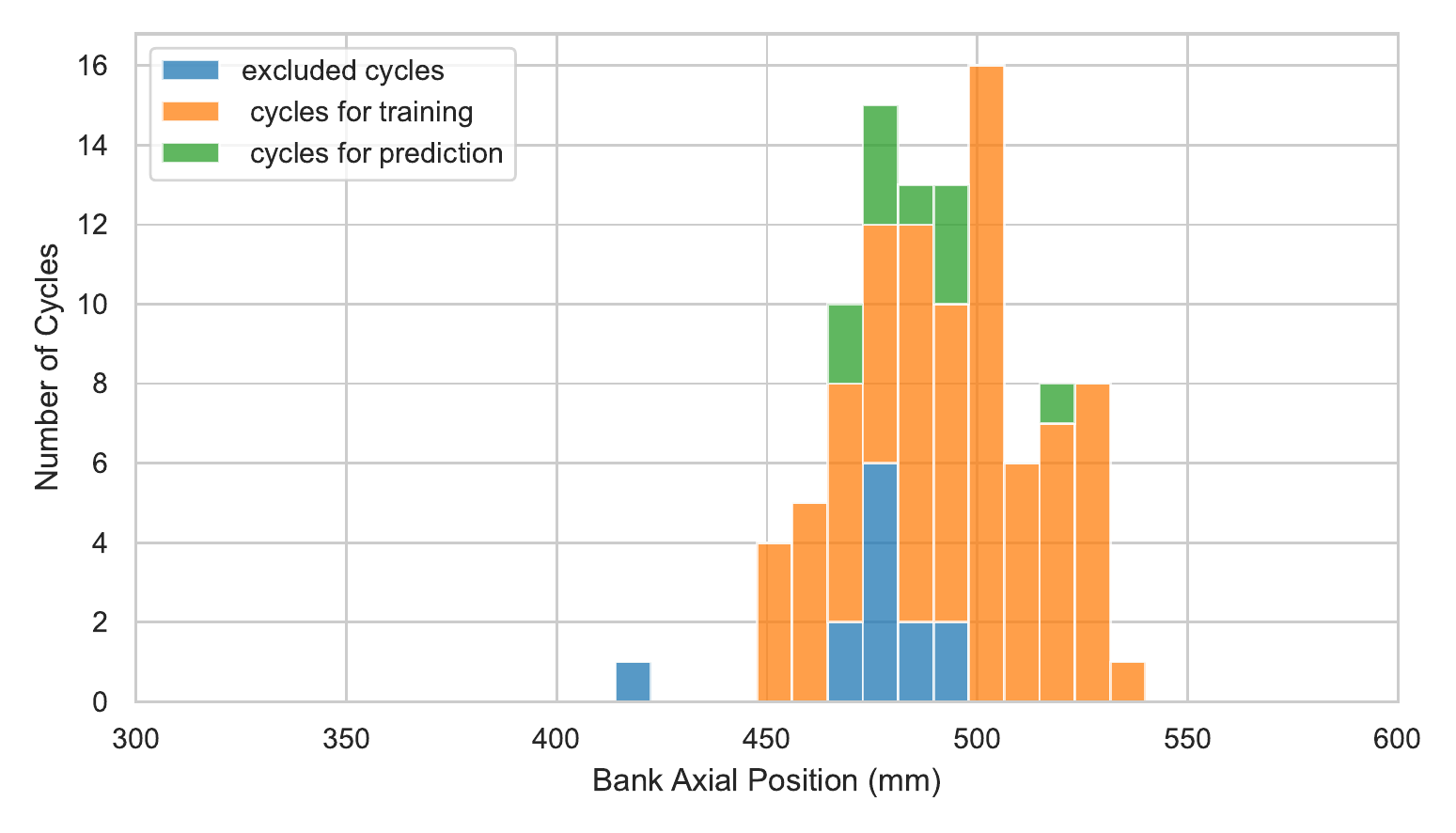}
  \caption[Distribution of control banks]{Distribution of control bank axial location for historical cycles. 
  }
  \label{fig:control-bank-distributions}
\end{figure}

In SAFARI-1, a fixed fuel loading pattern is used and the fuel management team and operators strive to employ this strategy for each cycle so as to avoid deviation from the normal operation and to minimize cycle-to-cycle variation of certain reactor parameters, including the neutron flux distribution. For this reason, the beginning of cycle (BOC) bank position does not vary significantly. This is illustrated by Figure~\ref{fig:control-bank-distributions}, which shows the distribution of control bank positions in the beginning of the historical cycles. As one may observe, the BOC position varied in a relatively narrow range from \SI{450}{\milli\meter} to \SI{550}{\milli\meter}, hence relatively small flux shape distortions may be expected.

\begin{figure}[!tb]
  \centering
  \includegraphics[width=0.45\linewidth]{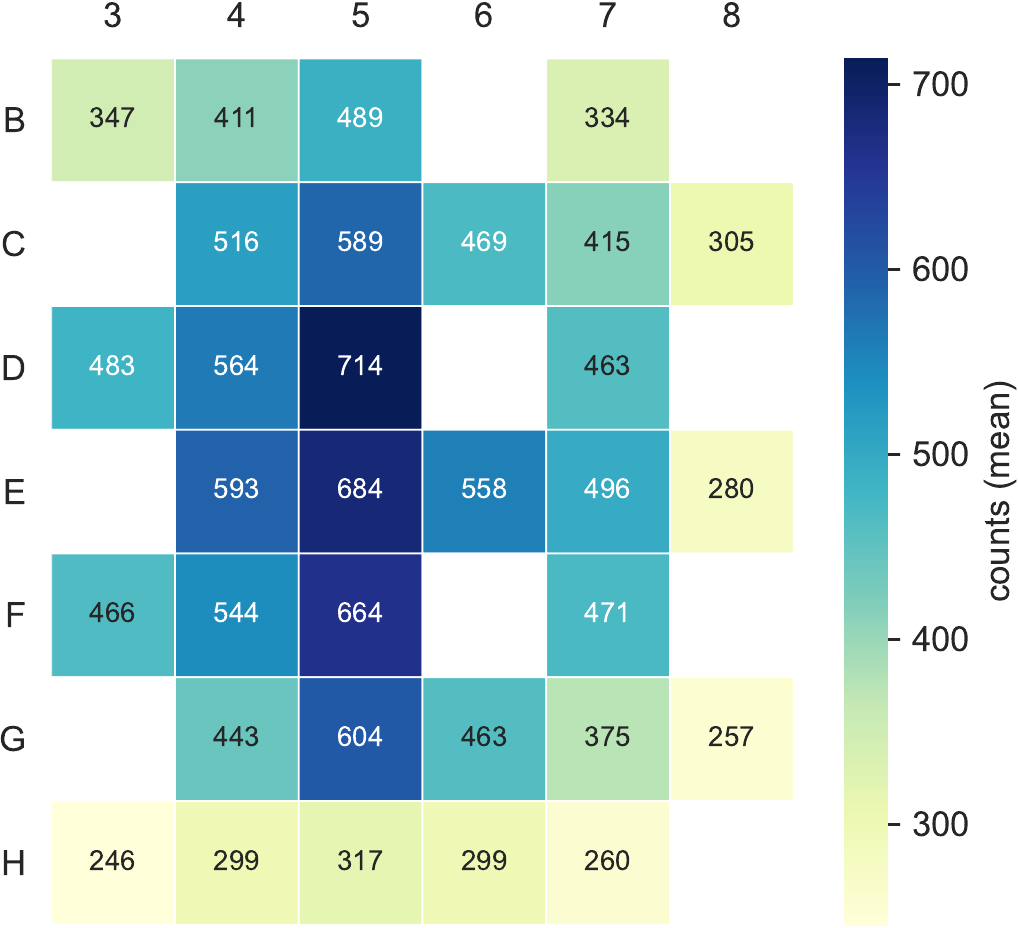}
  \caption{SAFARI-1  count distribution averaged over the axial height and over the cycles of interest.}
  \label{fig:count-distribution}
\end{figure}

\begin{figure}[!tb]
  \centering
  \includegraphics[width=0.45\linewidth]{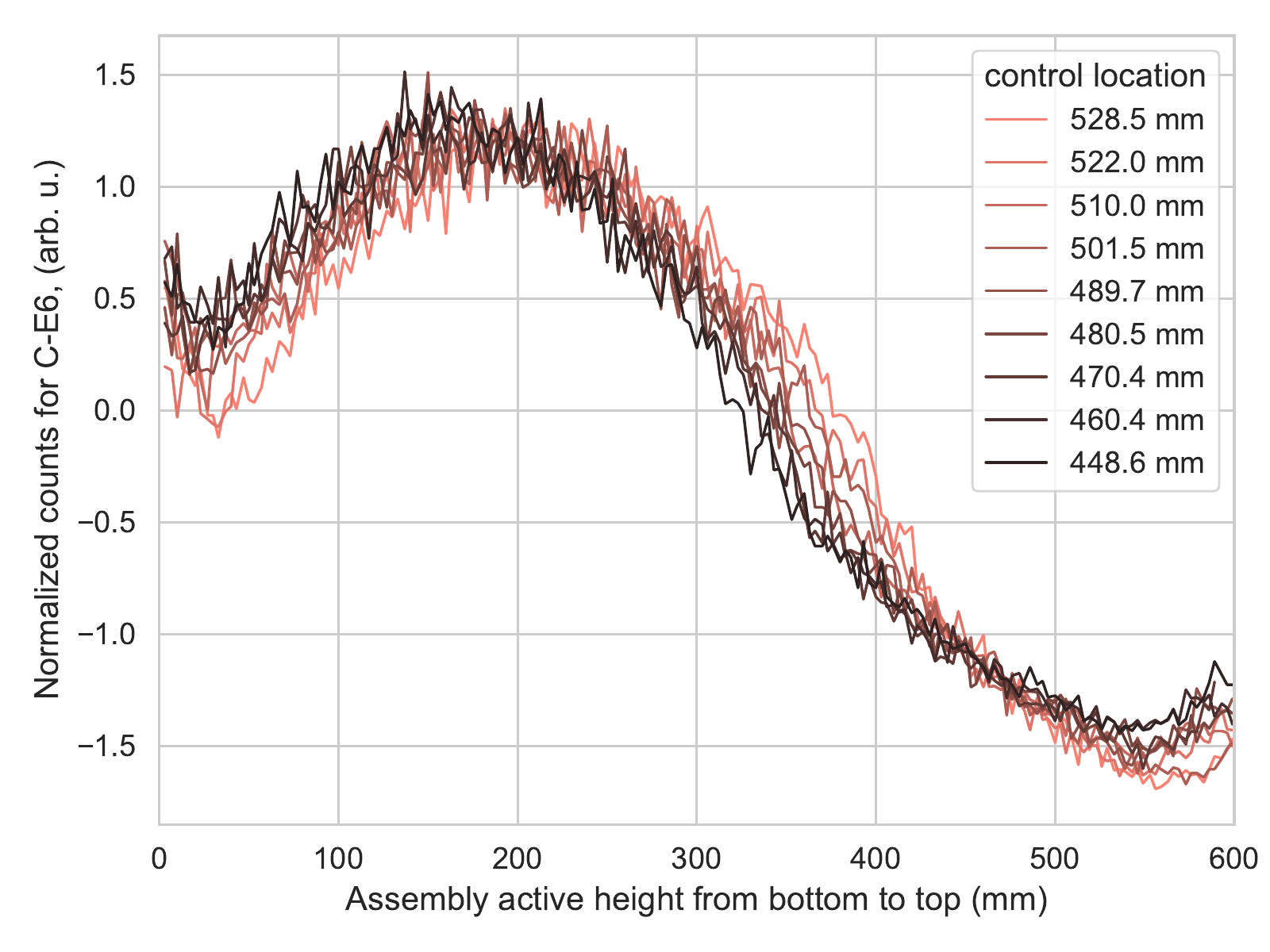}
  \includegraphics[width=0.45\linewidth]{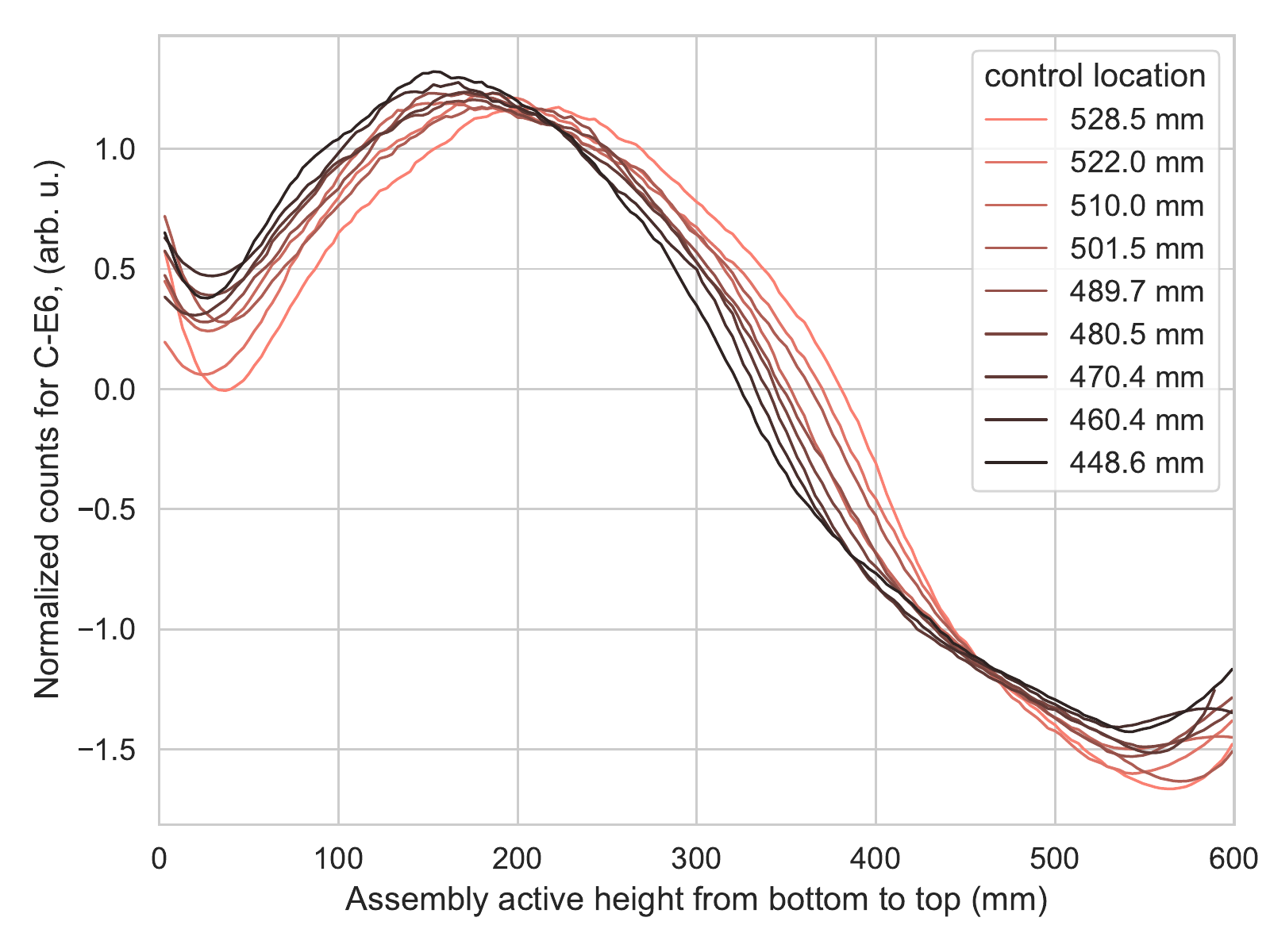}
  \includegraphics[width=0.45\linewidth]{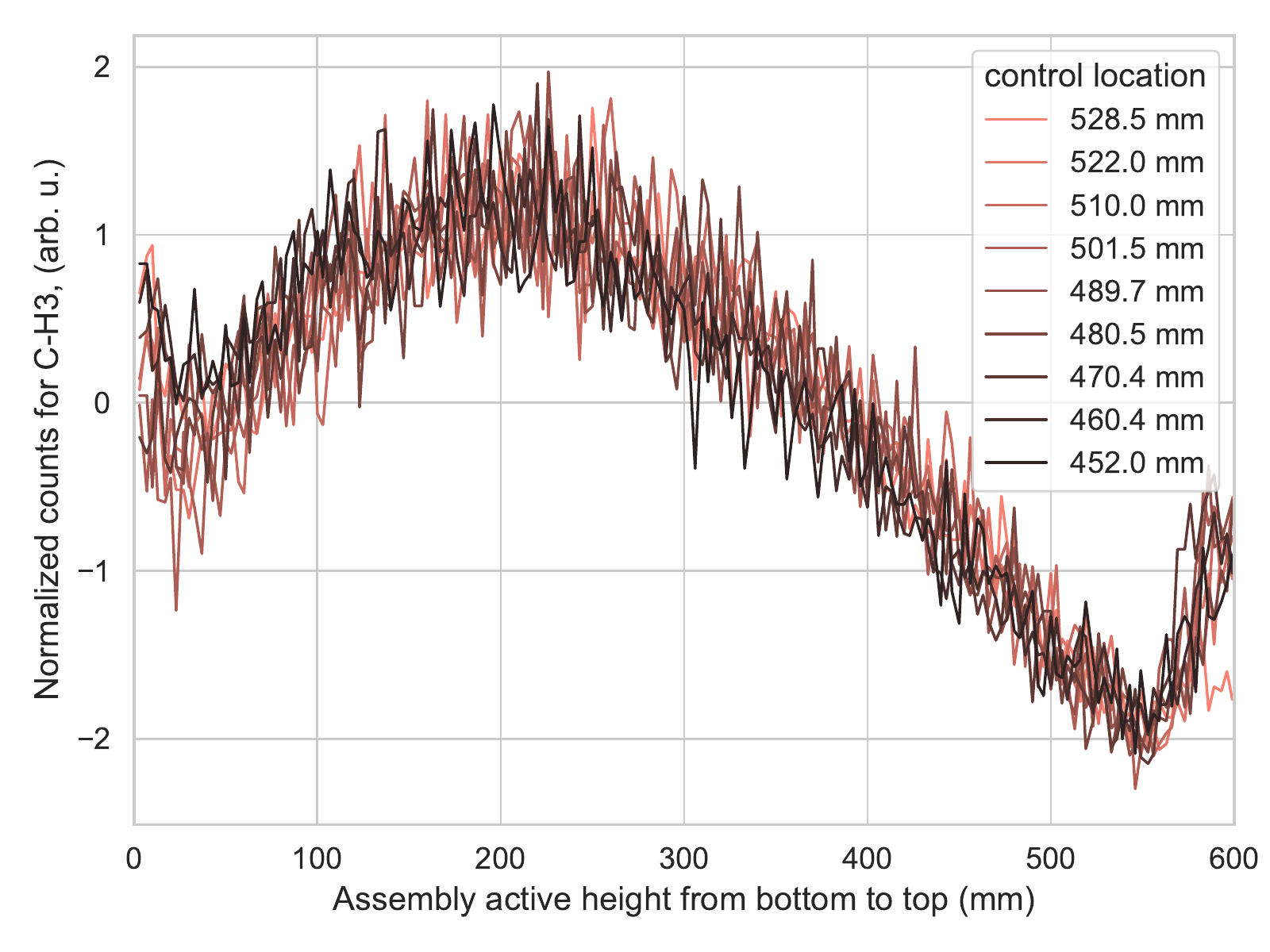}
  \includegraphics[width=0.45\linewidth]{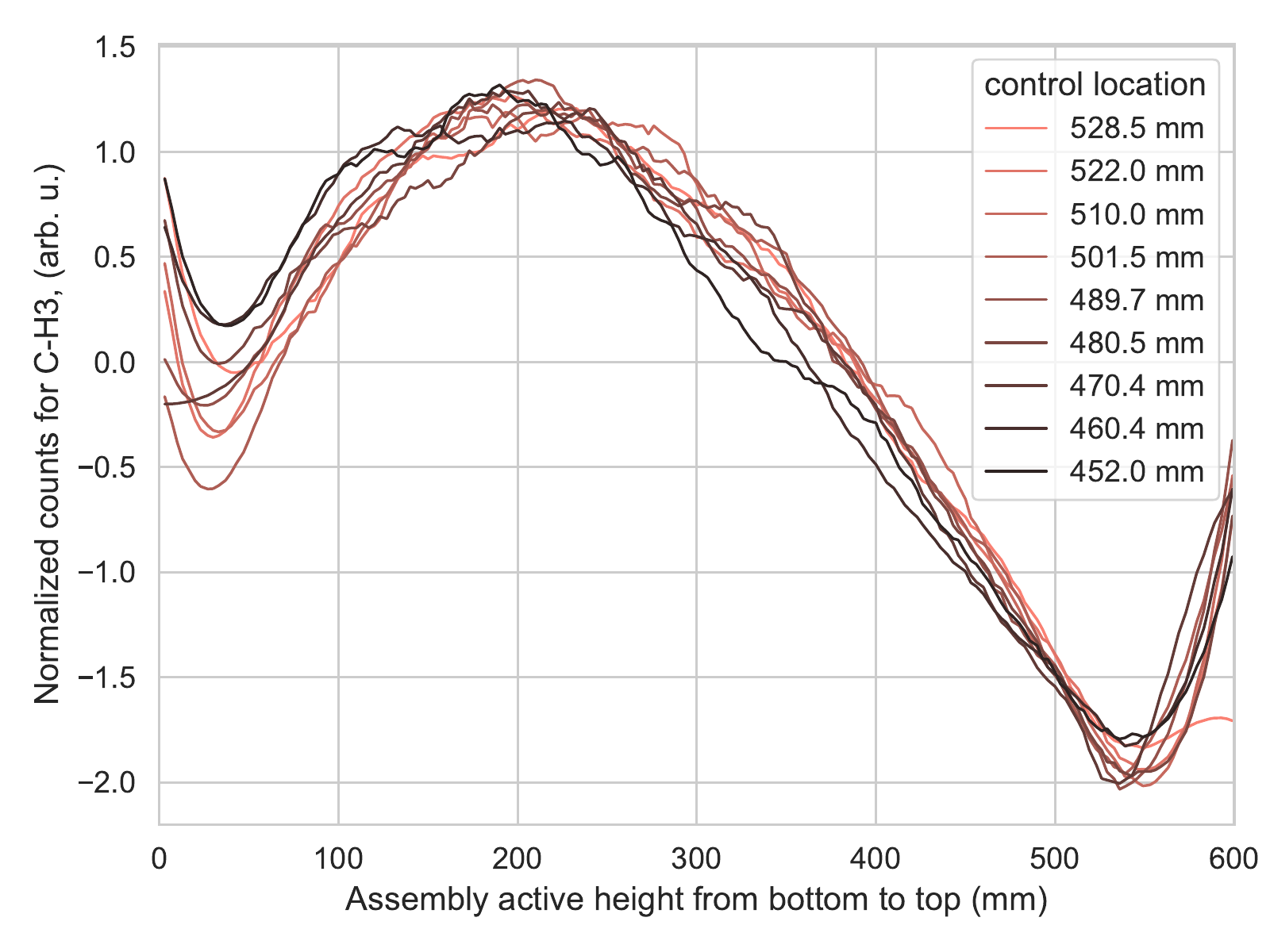}
  \caption{Normalized flux shapes for different control bank axial locations in core positions E6 and H3 before (left) and after (right) noise filtering.}
  \label{fig:c-e6}
\end{figure}

The second and the third factor are interlinked with a general tendency of the flux shape to be more sensitive to the bank axial position in the radially central region, where the control rods are located (readers are referred to Figure~\ref{figure:safaricore} for control rod placement). However, this observation is to a large extent obscured by the statistical noise in experimental data. The measured flux shapes are less noisy in the centre of the core, where the magnitude of the neutron flux is maximal and the statistical error is minimal. This is illustrated by Figures~\ref{fig:count-distribution} and \ref{fig:c-e6}. Figure~\ref{fig:count-distribution} shows an averaged (over the axial height and over the cycles of interest) count distribution in the core, thereby providing information required to estimate the magnitude of error.
In Figure~\ref{fig:c-e6}, nine normalized flux shapes  at core positions E6 and H3 are given for approximately equidistant axial bank locations covering the historical range. These core positions can be regarded as two limiting cases: centre of the core, vicinity of control rods and periphery of the core, far from control rods, respectively. 

Other fuel assemblies demonstrate a behaviour between these two, whereas control assemblies are considered separately. As one may observe in Figure~\ref{fig:c-e6}, in the presence of noise, the flux shape variation is barely observable for the central assembly and completely obscured by the noise for the peripheral assembly. For the sake of the analysis herein, the Savitzky-Golay noise filter \cite{Savitzky1964,Schafer2011} was therefore applied to the measured flux profiles, with optimally selected parameters of the filter obtained in \cite{moloko2021estimation}.
After applying the filtering procedure, it can be clearly seen that the assembly flux profiles are indeed sensitive to the bank movement for position E6, whereas the variation of the flux shape in position H3 is less pronounced and still being affected by the noise.
%

\subsection{Pre-processing of data for further analysis}
\label{subsec:preprocessing}

Data pre-processing is a procedure that converts raw data into a representation suitable for application through a sequence of operations. The objectives of data pre-processing include the reduction of the input space, obtaining smoother relationships, data normalization, noise reduction, and feature extraction \cite{Jain2005,Nawi2013,Yu2006,Schafer2011,Cleveland1979}. In the context of training a NN, it involves using techniques such as normalization and standardization to re-scale input and output data, prior to training.

The manual operation of the copper-wire measurement procedure, employed in the SAFARI-1 reactor, is associated with several challenges of a practical nature, which may have an impact on the quality of the measured data. These challenges may be related to the loading, irradiation, unloading and measuring activity of the copper wires after the irradiation.  As a result, the measured data may have miscellaneous deficiencies, which might be difficult for the operators to detect and locate. These deficiencies include, but are not limited to, statistical noise, missing measurement points, flux profiles shifted due to inadvertently displaced wires along the assembly active height, etc. For instance, over-exposure or under-exposure of wires which may lead to varying from cycle to cycle magnitudes of the measured axial flux shapes. The under-exposed wires may result in very noisy experimental measurement data (typically, the statistical noise is \SIrange{4}{6}{\percent}). 

The above-mentioned characteristics and deficiencies of experimental data necessitate pre-processing prior to using it for NN training. The measurement data, spanning 105 operational cycles from 2009 to 2021, was pre-processed for use in NNs training. The cycles or data with extremely low measurements (i.e.~with a maximum axial count value of $\leq 100$ in all the assemblies) were discarded from the dataset. 
The process yielded 86 cycles, available for training and testing of the NNs. 
Since wires are scanned and measured one at a time, the beginning of the scanning time differs per wire and thus influences the final measured counts. Further pre-processing of the 86 remaining cycles include correcting the scanned data to the same set time before measurements, by performing exponential decay corrections.
Prior to training the NN, it is important to note that the data with different scales often lead to the instability of NN, i.e. unscaled data could interfere with the training process and/or bias the trained NN models \cite{Nawi2013}. It is therefore necessary to ensure that the training data is normalized before training. In this work, the training data, i.e. input-output data, is normalized using $z$-score normalization \cite{Nawi2013}. Finally, in some studies the data was smoothed by applying the Savitzky-Golay filter as described in \cite{moloko2021estimation}.
Further details on the approach adopted in training the NNs for prediction of axial neutron flux profiles at a point, given the axial location and control bank position as inputs, are outlined in the subsequent sections.

\section{Theory}
\label{sec:theory}

This section provides a brief summary of the NN theory, key definitions, as well as training and UQ procedures as it pertains to this study.

\subsection{Artificial Neural Networks}
\label{subsec:ann}

An ANN consists of a large number of interconnected units, referred to as \emph{neurons} or \emph{nodes}, organized in \emph{layers}, that is, input layer, output layer, and one or more \emph{hidden} layer(s) between the input and output layers. In ANNs with one hidden layer, sometimes referred to as \emph{shallow NNs}, each input layer neuron is connected to every hidden layer neuron, which in turn is connected to the output layer neurons. Each connection is characterized by its strength, described with a parameter $w$ referred to as a \emph{weight}. 

An ANN can be regarded as a process where the input features go through a series of linear or non-linear transformations to predict the output quantities of interest.  Most ANNs do so by using an affine transformation, controlled by learnable parameters: weights, $\bm{W}$, and biases, $\bm{b}$, followed by a linear/non-linear transfer function referred to as \emph{activation function} $\phi(\cdot)$, such as hyperbolic tangent (tanh), rectified linear unit (ReLU), etc. \cite{Suzuki2011}. \emph{Architecture} of an ANN is described by the number of hidden layers, the number of neurons in each layer, the form of activation function, training algorithms, the learning rate and other pertinent parameters used in the NN \cite{Suzuki2011}. 

\emph{Training} an ANN refers to a procedure for determining a set of NN parameters, namely weights and biases, which map inputs to outputs in the best way. Training is performed by presenting the network with a set of known input-output pairs, $\mathcal{D} = \left\{ (\bm{x}_i, \bm{y}_i) \right\}_{i=1}^{N}$, where $N$ is the number of samples in the dataset. A training algorithm adjusts the NN parameters in such a way that the \emph{predicted} (computed) output $\hat{\bm{y}}_i=\hat{\bm{y}}(\bm{x_i})$ and the \emph{target} (actual) output $\bm{y}_i$ are close to each other. Training an NN involves using an optimization algorithm, which minimizes a \emph{cost function} that quantifies the accuracy of the approximation, e.g.~the Mean Squared Error (MSE): 
\begin{equation} \label{equation:mse}
    \text{MSE} = \frac{1}{N}\sum_{i=1}^N \Vert\hat{\bm{y}}_i-\bm{y}_i\Vert_2^2,
\end{equation}
or Mean Absolute Error (MAE)
\begin{equation} \label{equation:mae}
    \text{MAE} = \frac{1}{N}\sum_{i=1}^N \lvert\hat{\bm{y}}_i-\bm{y}_i\rvert.
\end{equation} 

\emph{Testing} of ANN models refers to a process of ensuring that the NN models are able to predict samples not used previously in training (referred to as a test dataset), while \emph{validation} guarantees the generalization of the model to samples not involved in the training process, in order to overcome the so-called \emph{over-fitting} problem.

\subsection{Feedforward Networks}

An NN design can be attributed to two broad classes based on the flow of signals across the network: in \emph{feedforward} networks, the connections between nodes do not form a loop, and in the \emph{recurrent} network, certain pathways are cycled. The feedforward networks are more widely used because they are simple from the viewpoint of structure and are easily analysed mathematically \cite[and references therein]{moloko2021estimation}. 

With the above description in mind, a regular feedforward ANN can be presented as follows:
\begin{equation}
    \begin{aligned}
        \bm{h}^{(0)} & = \bm{x}, \\
        \bm{h}^{(1)} & = \phi_1\left(\bm{W}^\top_1 \bm{h}^{(0)} + \bm{b}_1\right), \\
        & \vdots \\
        \bm{h}^{(L-1)} & = \phi_{L-1}\left(\bm{W}^\top_{L-1} \bm{h}^{(L-2)} + \bm{b}_{L-1}\right), \\
        \hat{\bm{y}} & = \bm{W}^\top_{L} \bm{h}^{(L-1)} + \bm{b}_{L},
    \end{aligned}
    \label{equation:architeture}
\end{equation}
where $L$ is the depth of the NN, including ($L-1$) hidden layers and one output layer, $\bm{W}_\ell$ is the matrix of weights and $\bm{b}_\ell$ is the vector of biases at hidden layer $\ell$, where $\ell=1,\ldots,L$, $\bm{h}^{(\ell)}$ is the vector of outputs from layer $\ell$, and $\hat{\bm{y}}$ is the prediction of the NN for input $\bm{x}$.

Training a feedforward network involves multiple forward and backward passes consecutively. In a forward pass, inputs are forwarded towards the output through multiple hidden layers of non-linearity and ultimately compare computed output with the actual output of the corresponding input. In the backward pass, the error derivatives with respect to the NN parameters are back propagated to adjust the weights in order to minimize the error in the output. The process continues multiple times until the desired improvement in prediction is obtained. 

In the training process, the back-propagation step is used to calculate the gradient of the cost function w.r.t. the NN parameters, which will further be used to update the parameters, using gradient descent. It is worth mentioning that the MSE is more sensitive to outliers compared to MAE, which is more robust with outliers \cite{PriceD2022}.


As discussed in \cite{Hornik1989}, standard multi-layer feedforward networks with just one hidden layer that contain a finite number of hidden neurons and arbitrary activation function, can fit any finite input-output mapping problem. The multi-layer feedforward networks can be used as a universal function approximator. 

\subsection{Deep Neural Networks}
\label{subsec:dnn}

\begin{figure}[!tb]
  \centering
  \includegraphics[width=.995\linewidth]{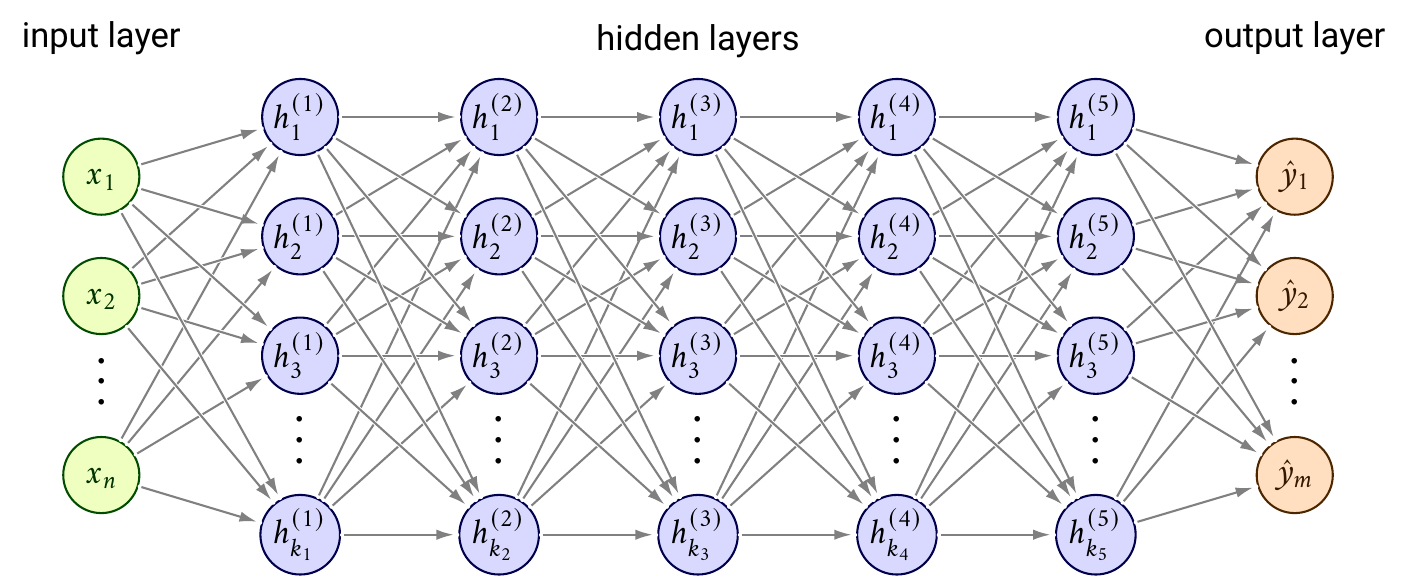}
  \caption{Example of a fully connected DNN with five hidden layers.}
  \label{figure:dnns}
\end{figure}

An ANN with multiple hidden layers is referred to as a \emph{Deep Neural Network} (DNN) or \emph{Deep Feedforward Neural Network} \cite{Lecun2015Deeplearning}. A typical DNN architecture is depicted in Figure~\ref{figure:dnns}, where an example of a fully connected DNN with one input layer with $n$ neurons, one output layer with $m$ neurons, and five hidden layers, each containing $k_{\ell}$ neurons ($\ell = 1,\ldots,5$) is shown. 

DNNs differ from shallow ANNs in that they can automatically learn representation from data (automatic feature extraction) without introducing hand-coded rules or human domain knowledge \cite{Lecun2015Deeplearning,Saptarshi2020Deeplearning}. They have been employed successfully and exhibited state-of-the-art performance in many fields such as, but not limited to, regression and classification problems. With the learnable parameters and activation function defined as in Section~\ref{subsec:ann}, equations~\eqref{equation:architeture} can be presented in the following compact form \cite{Gal2016Uncertainty,bachstein2019uncertainty}:
\begin{equation}
    \begin{aligned}
      \hat{\bm{y}}(\bm{x}) = \phi\left(\ldots\phi\left(\phi(\bm{x}\bm{W}_{1}+\bm{b}_{1})\bm{W}_{2}+\bm{b}_{2}\right)\ldots\right)\bm{W}_{L}+\bm{b}_{L},
    \end{aligned}\label{equation:dropout-I}
\end{equation}
where $\bm{b}_L$ is the bias vector and not scalar as would be in the case for an NN with one hidden layer. In this case, the MSE or MAE is minimized as the cost function.

In the practice of training DNNs, there are still two common issues: the potential vanishing or exploding gradient and over-fitting. The vanishing gradient is caused by the activation function, for which differential values are small, such as the tanh or sigmoid function. Many techniques have been introduced to combat these longstanding issues in training DNNs, with a varying degree of success. A good set of activation functions has been introduced to overcome the vanishing gradient problem, e.g.~the ReLU activation function, allowing the model to perform better, and the $L_1$ and $L_2$ regularization techniques alleviate over-fitting \cite{Srivastava2014dropout,Lecun2015Deeplearning,Saptarshi2020Deeplearning,Liu2018DNN}. It is important to note that the standard DNN only predicts deterministic outputs given an input, without providing the approximation/prediction uncertainties. A discussion on the estimation of model uncertainty, using different methods, is provided in Sections~\ref{subsec:mcd} and \ref{subsec:bnn}.

\subsection{Monte Carlo Dropout}
\label{subsec:mcd}

\begin{figure}[!tb]
	\centering
	\includegraphics[width=.95\linewidth]{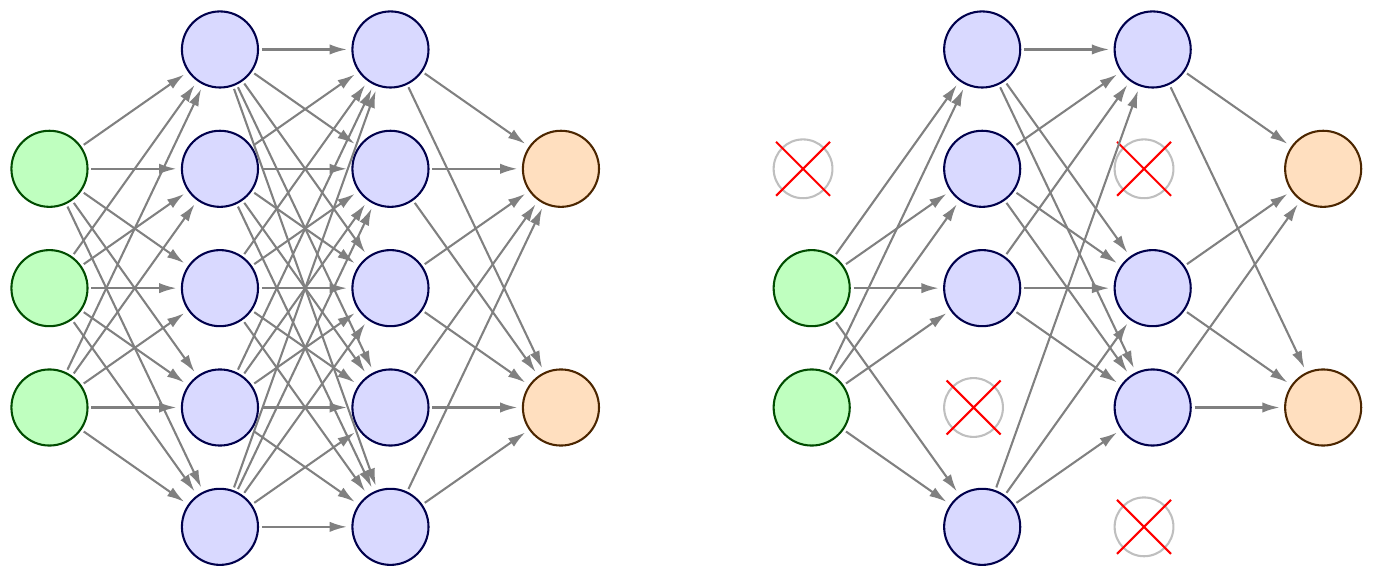}
	\caption{Illustration for the dropout technique: dropping out nodes in a given NN.}
	\label{figure:regular-vs-dropout}
\end{figure}

Dropout \cite{Srivastava2014dropout,Saptarshi2020Deeplearning} is an effective technique that has been widely used to overcome the over-fitting problem in DNNs.  Dropping a unit out means temporarily removing it from the network, along with all its incoming and outgoing connections, as is illustrated in Figure~\ref{figure:regular-vs-dropout}. During training, certain nodes (i.e. neurons) in hidden layers are randomly ignored or ``dropped out''. This has the effect of making the layer look-like and be treated-like a layer with a different number of nodes and connectivity to the prior layer. Dropout for regularization is used during DNN training but not in prediction~\cite{Srivastava2014dropout}.

Mathematically, the dropout can be expressed as:
\begin{equation}
    \begin{aligned}
      \hat{\bm{y}}(\bm{x}) = \phi\left(\cdots\phi\left(\phi(\bm{x}{\bm{Z}_{1}}\bm{W}_{1}+\bm{b}_{1}){\bm{Z}_{2}}\bm{W}_{2}+\bm{b}_{2}\right)\cdots\right){\bm{Z}_{L}}\bm{W}_{L}+\bm{b}_{L},
    \end{aligned}\label{equation:dropout-II}
\end{equation}
where $\bm{Z_}\ell$ denote random diagonal matrices
\begin{equation}
    \begin{aligned}
      \bm{Z_}\ell = \text{diag}\left(z_{\ell,1},\ldots,z_{\ell,k_{\ell-1}}\right),
    \end{aligned}
\end{equation}
in which $z_{\ell,j}\sim\text{Bernoulli}(p_{\ell})$ are independent Bernoulli random variables with some probability $p_\ell$ for $\ell=1,\dots,L$ and $j = 1,\ldots,k_{\ell-1}$.

The objective function to be minimized, using $L_2$ regularization with some weight decay $\lambda$, is given by:
\begin{equation}
    \begin{aligned}
          \mathcal{L}_{\text{dropout}}
          = \frac{1}{N}\sum_{i=1}^{N}
          {\left\Vert \hat{\bm{y}}_{i}-\bm{y}_{i}\right\Vert ^{2}}
          +\lambda\sum_{\ell=1}^{L}\left(\left\Vert \bm{W}_{\ell}\right\Vert _{2}^{2}+\left\Vert \bm{b}_{\ell}\right\Vert _{2}^{2}\right).
    \end{aligned}\label{equation:loss_dropout}
\end{equation}

The dropout technique samples binary variables for each input value and every NN unit in each layer (except for the output layer). The binary variables take values of ``1'' with probability $p_\ell$ for layer $\ell$, except in cases where it takes a value of ``0'', in this case it is dropped for a given input value. The same values are used in a backward pass to update the parameters \cite{Srivastava2014dropout,Saptarshi2020Deeplearning}.

Monte Carlo Dropout (MCD) \cite{gal2016dropout} is a technique to quantify uncertainties in DNNs. In MCD, dropout is used to introduce randomness to both the training and prediction processes. Once trained, the NN can be evaluated for the same input multiple times, also with dropout of the hidden neurons, resulting in a collection of predictions that can be used to estimate the uncertainties in the predictions. Empirical distributions over the outputs can be derived, which can be used to obtain, for example, mean values and confidence measures in terms of the distributional variance. The empirical variance (i.e. prediction uncertainty) is expected to be low in the case of abundant training data, since all NN subsets (smaller NN after dropout) had the opportunity to learn in these training domains. However, in domains with little or no training data to learn from, the NN behaviour is not controllable, so we expect a high variance among the different NN subsets. In extrapolation (generalization) regions, the uncertainty will be large, while in the interpolation (training) regions, the uncertainty will be small. This is exactly what we want for uncertainty estimation purposes. By adding the dropout layers $\bm{Z}_1,\ldots,\bm{Z}_L$
and introducing the scaling factors between layers in the regular ANN, we can express the MCD process as follows:
\begin{equation}
    \begin{aligned}
  \hat{\bm{y}}(\bm{x}) = \sqrt{\frac{1}{k_{L-1}}}\phi\left(...\sqrt{\frac{1}{k_{2}}}\phi\left(\sqrt{\frac{1}{k_{1}}}\phi\left(\bm{x}{\bm{Z}_{1}}\bm{W}_{1}+\bm{b}_{1}\right){\bm{Z}_{2}}\bm{W}_{2}+\bm{b}_{2}\right)...\right){\bm{Z}_{L}}\bm{W}_{L}+\bm{b}_{L},
      \end{aligned}
      \label{equation:mcd}
\end{equation}%
where $k_\ell$ is the number of nodes in the $\ell$-th hidden layer and $\bm{Z}_\ell$ is the dropout matrix before the $\ell$-th layer with some Bernoulli probability $p\in\left(0, 1\right)$, which is the same for all dropout layers. In MCD, the loss function of Eq.~\eqref{equation:loss_dropout} is scaled by the probability $p$ and is given by:
\begin{equation}
    \begin{aligned}
          \mathcal{L}_{\text{dropout}}
          = \frac{1}{N}\sum_{i=1}^{N}
          {\left\Vert \hat{\bm{y}}_{i}-\bm{y}_{i}\right\Vert ^{2}}
          +\lambda\sum_{\ell=1}^{L}\left(p\left\Vert \bm{W}_{\ell}\right\Vert _{2}^{2}+\left\Vert \bm{b}_{\ell}\right\Vert _{2}^{2}\right).
    \end{aligned}\label{equation:mcdloss}
\end{equation}

In MCD, the mean prediction is estimated by averaging over $T$ forward passes while the uncertainty can be estimated in terms of the standard deviation:

\begin{equation}
      \begin{aligned}
        \hat{\bm{\mu}}(\bm{x}) =\frac{1}{T}\sum_{t=1}^{T}\hat{\bm{y}}^{(t)}(\bm{x}),\quad
        \hat{\sigma}^{2}(\bm{x})  =\frac{1}{T-1}\sum_{t=1}^{T}\left(\hat{\bm{y}}^{(t)}(\bm{x})-\hat{\bm{\mu}}(\bm{x})\right)^{2}.
      \end{aligned}
\end{equation}


\subsection{Bayesian Neural Networks}
\label{subsec:bnn}

\begin{figure}[!tb]
	\centering
	\includegraphics[width=.995\linewidth]{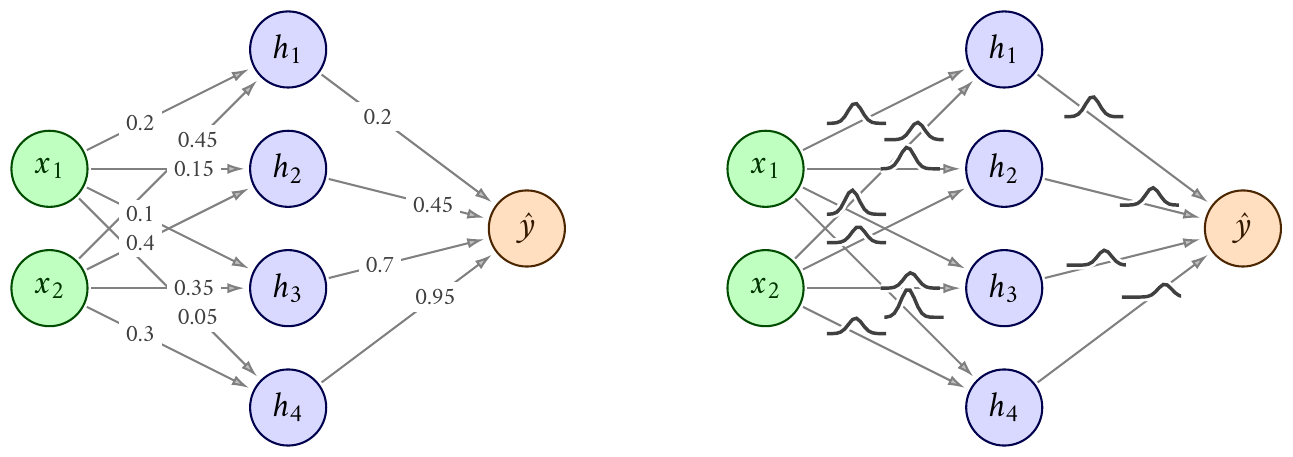}
    \caption{Regular DNN (left) and BNN (right).}
	\label{figure:BNN-Illustration}
\end{figure}

When training a regular ANN, the weights and bias are updated with stochastic gradient descent using the derivatives of the cost function w.r.t. these parameters. The training process stops when the cost function reaches a sufficiently low value or stops decreasing. The weights and bias in a trained NN model have deterministic values. Therefore, when the trained NN model predicts at a new input, a deterministic output is obtained without uncertainty. A BNN \cite{goan2020bayesian} is an NN that assigns uncertain distributions over its parameters (weights and bias), as is illustrated in Figure~\ref{figure:BNN-Illustration}. Bayesian inference will be used to determine the NN parameters. They are first assigned prior distributions, given the training data, the posterior distributions of the NN parameters will be computed. This Bayesian inference process is accomplished in the training step. In the prediction step, every time the trained BNN is evaluated on a specific input, the predictions can be different, which are used to quantify the predictive uncertainty.

As weight parameters of a BNN are represented by probability distributions over possible values, rather than having a single fixed value, instead of training a single network, the BNN essentially trains an ensemble of networks, where each NN has its weights drawn from a learnt probability distribution. Since exact Bayesian inference is computationally intractable for BNNs, a variety of approximations have been developed. The most popular approaches are Markov Chain Monte Carlo (MCMC) methods \cite{neal2012bayesian} and variational Bayesian (VB) methods, also called variational inference (VI) \cite{blei2017variational,tzikas2008variational}. MCMC methods try to approximate the posterior distribution of the parameters using sampling, which has slow convergence, while VB/VI methods shift from sampling to optimization, which can be much faster for complex models and larger datasets.

The method, that employs VI for training BNN weights, is referred to as ``Bayes by Backprop'' in \cite{blundell2015weight}. In order to better explain this method, we can first interpret an NN as a probabilistic model: $P (\bm{y} | \bm{x}, \bm{w})$, where $\bm{x} \in \mathbb{R}^n$ is the input vector, $\bm{w}$ is a set of parameters such as weights, and $\bm{y}$ is the output. The NN maps input $\bm{x}$ onto the parameters of a distribution on $\bm{y}$ by several successive layers of linear transformation (given by $\bm{w}$) interleaved with element-wise non-linear transforms (with activation functions). The weights can be learnt by Maximum Likelihood Estimation (MLE), i.e. given a set of training data $\mathcal{D} = \left\{ (\bm{x}_i, \bm{y}_i) \right\}_{i=1}^{N}$, the MLE of the weights $\bm{w}^{\text{MLE}}$ are defined as:
\begin{equation}
	\bm{w}^{\text{MLE}}  =  \argmaxA_{\bm{w}}  \log P (\mathcal{D} | \bm{w})   =  \argmaxA_{\bm{w}}  \sum_{i=1}^{N} \log P (\bm{y}_i | \bm{x}_i, \bm{w}).
\end{equation}

The training process is done by employing gradient descent (with back-propagation), where we assume that $\log P (\mathcal{D} | \bm{w})$ is differentiable in $\bm{w}$. MLE returns deterministic estimates of $\bm{w}$, as in the training of a regular ANN. In contrast, BNN targets at posterior distributions of $\bm{w}$. BNN calculates the posterior distribution $P (\bm{w} | \mathcal{D})$ of the weights, given the training data, based on the Bayes' rule:
\begin{equation}
	P (\bm{w} | \mathcal{D})  =  \frac{P (\bm{w}, \mathcal{D})}{P (\mathcal{D})}  =  \frac{P (\mathcal{D} | \bm{w}) \cdot P (\bm{w})}{P (\mathcal{D})},
\end{equation}
where $P (\bm{w})$ is the prior distribution for $\bm{w}$, $P (\mathcal{D} | \bm{w})$ is the likelihood function, and $P (\bm{w} | \mathcal{D})$ is the posterior distribution for $\bm{w}$. Prior and posterior represent our knowledge of $\bm{w}$ before and after observing the training data $\mathcal{D}$, respectively. When making predictions at a test data $\hat{\bm{x}}$, the predictive distribution of the output $\hat{\bm{y}}$ is given by:
\begin{equation}
	P (\hat{\bm{y}} | \hat{\bm{x}})  =  \mathbb{E}_{P (\bm{w} | \mathcal{D})} \left[  P (\hat{\bm{y}} | \hat{\bm{x}}, \bm{w})  \right],
\end{equation}
where the expectation operator $\mathbb{E}_{P (\bm{w} | \mathcal{D})}$ means that we need to integrate over $P (\bm{w} | \mathcal{D})$.

Unfortunately, such an expectation operation is intractable for ANNs of any practical size, due to the large number of parameters, as well as the difficulty to perform exact integration. This is why we need to find a variational approximation to  $P (\bm{w} | \mathcal{D})$. VI methods are a family of techniques for approximating intractable integrals arising in Bayesian inference and machine learning. They are typically used in complex statistical models, consisting of observed variables (``data'') as well as unobserved/latent variables. VI is used to approximate complex posterior probabilities that are difficult to evaluate directly as an alternative strategy to MCMC sampling. In particular, whereas MCMC provides a numerical approximation to the posterior using a sequence of samples, VI provides a locally-optimal, exact analytical solution to an approximation of the posterior with greater speed and comparable accuracy. The general idea of VI is to use a variational distribution $Q (\bm{w} | \bm{\theta})$, which is a family of distributions controlled by parameters $\bm{\theta}$, to approximate $P (\bm{w} | \mathcal{D})$. We seek to find an optimal value $\bm{\theta}^{*}$ of $\bm{\theta}$ that can minimize the Kullback-Leibler divergence $D_{\rm KL}$ from $P (\bm{w} | \mathcal{D})$ to $Q (\bm{w} | \bm{\theta})$:
\begin{align}
	\bm{\theta}^{*} \nonumber
	&=  \argminA_{\bm{\theta}}  D_{\text{KL}} \left( Q (\bm{w} | \bm{\theta}) \Vert P (\bm{w} | \mathcal{D}) \right) 
	=  \argminA_{\bm{\theta}}  \int  Q (\bm{w} | \bm{\theta}) \log  \frac{Q (\bm{w} | \bm{\theta})}{P (\bm{w} | \mathcal{D})}   d \bm{w}    \\  \nonumber
	&=  \argminA_{\bm{\theta}}  \int  Q (\bm{w} | \bm{\theta}) \log  \frac{Q (\bm{w} | \bm{\theta})}{P (\mathcal{D} | \bm{w}) \cdot P (\bm{w})}  d \bm{w}    \\
	&=  \argminA_{\bm{\theta}}  \left(  D_{\text{KL}} \left( Q (\bm{w} | \bm{\theta}) \Vert P (\bm{w}) \right)  -  \mathbb{E}_{Q (\bm{w} | \bm{\theta})} \left[  \log P (\mathcal{D} | \bm{w})  \right]  \right).
	\label{eq:optimal-params-theta}
\end{align}
The resulting cost/loss function, denoted as $\mathcal{F} (\mathcal{D}, \bm{\theta})$, is known as the \emph{variational free energy}, or \emph{expected lower bound} \cite[and references therein]{blundell2015weight}:
\begin{equation}
	\mathcal{F} (\mathcal{D}, \bm{\theta})  =  D_{\text{KL}} \left( Q (\bm{w} | \bm{\theta}) \Vert P (\bm{w}) \right)  -  \mathbb{E}_{Q (\bm{w} | \bm{\theta})} \left[  \log P (\mathcal{D} | \bm{w})  \right].
\end{equation}
The first part of $\mathcal{F} (\mathcal{D}, \bm{\theta})$ is prior-dependent and referred to as the complexity cost. The second part is data-dependent and referred to as the likelihood cost. The cost function embodies a trade-off between matching the data $\mathcal{D}$ (likelihood) and satisfying the simplicity prior $P (\bm{w})$. In order to evaluate the cost function $\mathcal{F} (\mathcal{D}, \bm{\theta})$, it will be re-organized in the form of expectation and evaluated using Monte Carlo sampling of $\bm{w}$ from the variational distribution $Q (\bm{w} | \bm{\theta})$. 
Further detail on this can be found in \cite{blundell2015weight}.

\section{Design and training of Neural Networks}
\label{sec:methodology}

\subsection{Training data}
\label{subsec:traning_data}

Data from 86 cycles, pre-processed as discussed in Section~\ref{subsec:preprocessing}, were subdivided into 76 cycles for training NN models and the remaining 10 cycles were reserved for evaluating the generalization and prediction capability of the trained models. The training data from 76 cycles, consisting of \num{13860} samples) were randomly partitioned into \SI{64}{\percent} for training, \SI{20}{\percent} for testing and \SI{16}{\percent} for validation. Data from 10 cycles, not used either in model training or in validation, will be referred to as ``prediction dataset'' or ``prediction cycles'' in this work and can be regarded as an additional test set, though it serves a different purpose, as explained above.  These datasets were used in conjunction with several NN training methodologies employed in this study and discussed in subsequent subsections.

\subsection{Design and training of DNNs with Keras}
\label{subsec:design_dnn}

\begin{figure}[!tb]
  \centering
  \includegraphics[height=.385\linewidth]{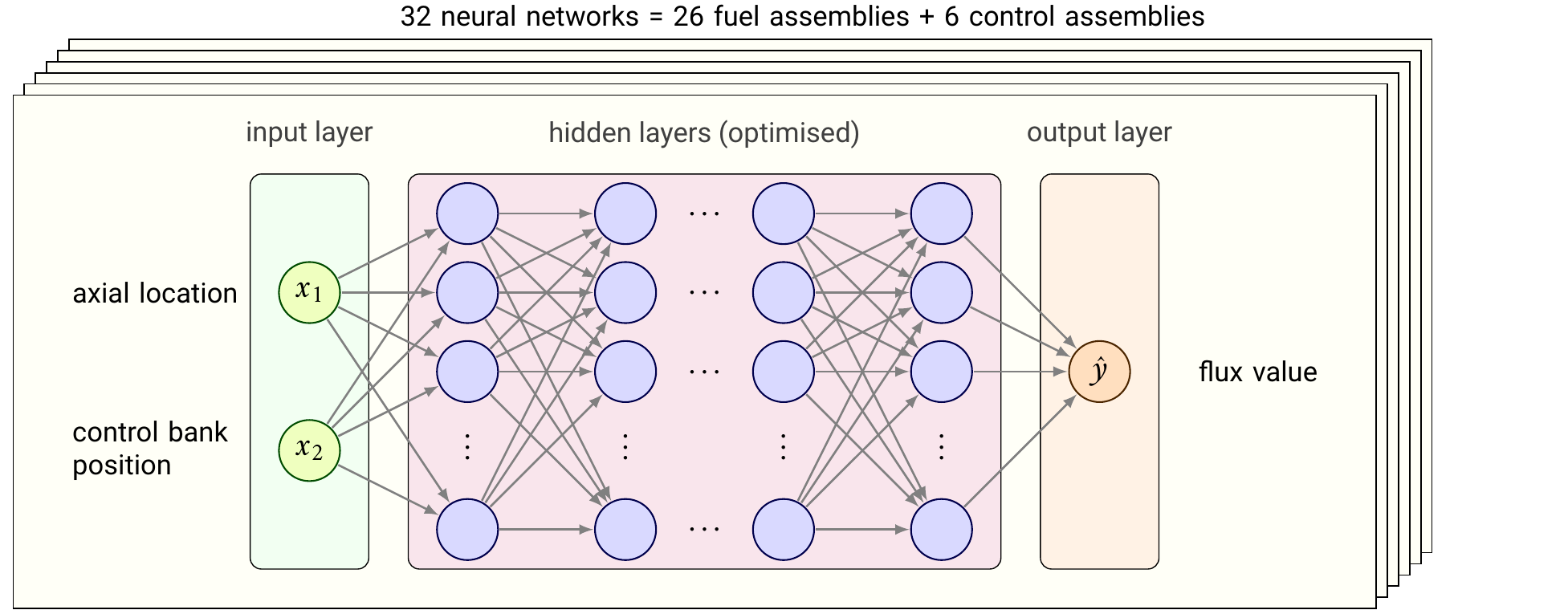}
  \caption{DNN architecture for prediction of axial neutron flux \cite{moloko2022UQ}.}
  \label{fig:fig-DNN-architecture}
\end{figure}

Our starting point and baseline is training a set of DNN models for the axial neutron flux shape. In this study, for each of the 32 assemblies (26 fuel- and 6 control follower assemblies), we constructed a DNN model, which  consists of an input layer with two neurons, corresponding to the control bank position and the axial location along the wires, one output layer with a single neuron representing the axial neutron flux value and several hidden layers. The number of hidden layers and number of neurons per layer were subject to optimization. The DNN architecture as adopted in the study is depicted in Figure~\ref{fig:fig-DNN-architecture}.   

All the analyses and results of this case are conducted using Python under the Keras deep learning package with TensorFlow backend \cite{chollet2015keras}. The DNN models make use of the \texttt{ReLU} activation function in the hidden layers and the \texttt{Adam} optimizer was used in training, as default. Four hyperparameters were selected for optimization, that is, the number of hidden layers, number of hidden units per layer, the learning rate, and batch size. Hyperparameter optimization was performed using Random Search and Grid Search algorithms implementation in the NeuroEvolution Optimization with Reinforcement Learning (NEORL) framework \cite{neorl2021}. NEORL has the advantage that most of its algorithms, including Random Search and Grid Search, supports parallel or multi-processing optimization. 
 
%

As discussed in \cite{BergstraB2012}, chances of finding the optimal hyperparameters are comparatively higher in Random Search than in Grid Search. With this in mind, and as a first step, the Random Search was used to narrow the choices of the important hyperparameters, that is, the number of hidden layers, and the number of hidden units per layer, while fixing the values of the learning rate and batch size. As a second and final step, the Grid Search is executed with more refined and narrower grids of the learning rate and batch size. The number of hidden layers and hidden units per layer was fixed and used as obtained from Random Search. Since the experimental measurements data is noisy (i.e. contains outliers), MAE was minimized instead of MSE.

To further improve the performance of Keras DNN models and reduce the over-fitting, a Keras regularization technique, namely \texttt{EarlyStopping}, was used. This technique monitors the performance of the model for every epoch on a held-out validation set during the training, and terminate the training conditional on the validation performance (i.e. when the validation error starts increasing from its minimum value). In addition, and in order to further improve the performance of the produced DNNs, the \texttt{ReduceLROnPlateau} callback was used. This ensures that the learning rate is reduced when a metric (e.g., validation error) has stopped improving. As a final step, the best (optimized) DNN model for each of the assemblies were used for the prediction of the axial neutron flux and results were compared to copper-wire experimental measurements. The  discussions are provided in Subsection~\ref{subsec: dnn_results}. 

During the process of hyperparameter optimization, it was observed that the DNN models are sensitive to the choice of hyperparameters and therefore it is important to ensure that the selected values are close to optimal as well as for the trained models. In practice, it turned out that the optimization of hyperparameters is a challenging task due to a combination of factors, including the noisy data, large number of hyperparameters, limited computational resources at our disposal, etc. One possible way to check the quality of obtained DNNs is to cross-validate them with models produced independently, preferably with different tools and methods. To this end, and in order to demonstrate the robustness of the Keras DNN models, their predictions were benchmarked with those obtained from the MATLAB Neural Network models produced in \cite{moloko2021estimation}. A brief summary of the MATLAB NN model design and training is provided in Subsection~\ref{subsec:matlab_ann}.

\subsection{Design and training of ANNs with MATLAB}
\label{subsec:matlab_ann}

\begin{figure}[!tb]
  \centering
  \includegraphics[height=.385\linewidth]{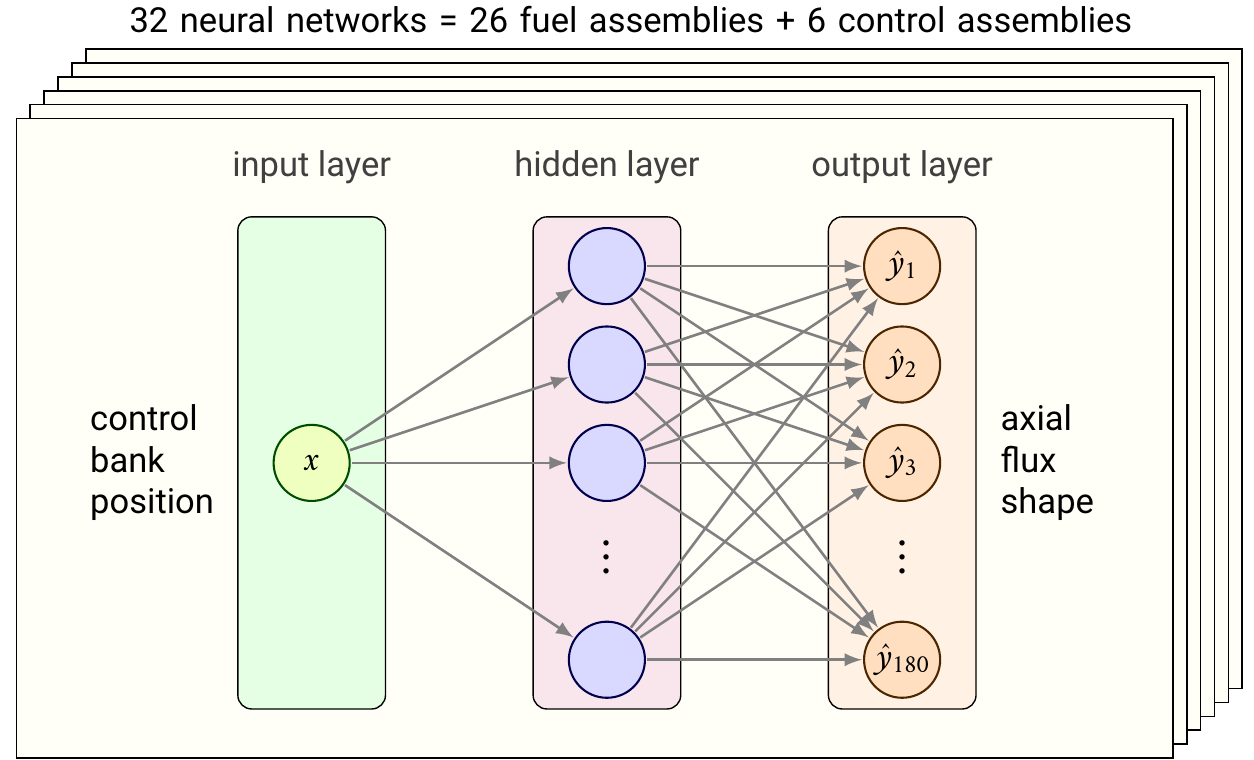}
  \caption{Shallow NN architecture for prediction of axial neutron flux \cite{moloko2021estimation} in MATLAB.}
  \label{figure:Matlab-SNN-architecture}
\end{figure}

The MATLAB Neural Network Toolbox \cite{MATLAB2017a}, that is, the NN fitting tool (\texttt{nftool}), was utilized in constructing and training the NN models in \cite{moloko2021estimation}. The MATLAB-trained three-layered feedforward shallow NN, consisting of one input layer with a single neuron (control bank position), one hidden layer and one output layer with 180 neurons (corresponding to 180 axial neutron flux measurements), as depicted in Figure~\ref{figure:Matlab-SNN-architecture}, was used to benchmark the DNN models. 

As with DNNs, 32 models (i.e. 26 fuel- and 6 control assemblies) were produced. A manual parametric study with the goal to find an optimal number of hidden neurons for each of the 32 assembly NNs was conducted. To this end, the number of neurons was varied from 1 to 30 and the best one, i.e. minimizing the MSE, was chosen for each NN.
%

The MATLAB NN models were trained using copper-wire measurement data from 47 cycles (only complete and good-quality data were manually selected and used for training). The input-output were normalized using $z$-score normalization and noise in the data was reduced using the Savitzky-Golay filter. The hyperbolic tangent (tanh) activation function was used in the hidden layer and a linear activation function in the output layer. The MSE cost function was minimized. Two algorithms were used in the training of the NN models, namely, the Levenberg-Marquardt and Bayesian Regularization algorithms, and it was concluded that the latter outperformed the former. Therefore, the results based on the Bayesian regularization algorithm are used for the comparison. Further details of the model can be found in \cite{moloko2021estimation}.

\subsection{Design and training of MCD and BNN VI models}
\label{subsec:mcd_bnn}

Similar to the Keras DNN models, the MCD and BNN VI implementations use \texttt{EarlyStopping} and \texttt{ReduceLROnPlateau} regularization techniques. Furthermore, the BNN VI uses the same NN architecture and hyperparameters as those of the Keras DNN models. On the other hand, the MCD models employ a bigger NN architecture than that of the DNNs. For illustration purposes, the NN architecture and their associated hyperparameters for fuel assembly E6 are summarized in Table~\ref{tab:hyperparameters}. Note that the MCD architecture and hyperparameters, were obtained from our prior experience (based on  manual ``trial and error'' analysis) and this model is fixed and used for all 32 assembly-wise NNs. In building the NN UQ model in MCD and BNN VI, the Keras TensorFlow Probability library was used. TensorFlow Probability provides an integration of probabilistic methods with deep networks. 
%
%
The MCD model makes use of the regular \texttt{Dense} and \texttt{Dropout} layers, with a dropout rate of 0.2. The BNN VI model makes use of a \texttt{DenseFlipout} layer, which accounts for the implementation of the Bayesian variational inference analogue to a standard dense layer, by assuming that the weights and/or the biases are drawn from statistical distributions. The layer implements a stochastic forward pass, by default, via sampling from the weights and biases posterior.  The outputs (i.e. $\mu$ and $\sigma$) of the MCD and BNN VI models are connected to the distribution via the TensorFlow Probability \texttt{DistributionLambda} layer. The Negative Log-Likelihood function of the normal distribution, the predicted mean ($\mu$) and uncertainty ($\sigma$) were sampled from a heteroscedastic Gaussian distribution.
 
The predictive mean was estimated by averaging over \num{20000} iterations for both BNN VI and MCD. The uncertainty is estimated in terms of the standard deviation. The chosen number of iterations was necessary to ensure convergence in the predictions of both the mean and the standard deviation.

\section{Results and discussion}
\label{sec:results-discussion}

The results of this section are divided into three sections. Section~\ref{subsec:hyperparameters} summarizes the results of the hyperparameter optimization, Section~\ref{subsec: dnn_results} presents the comparison of Keras DNN and MATLAB ANN and Section~\ref{subsec:UQ} is devoted to the quantification of uncertainties associated with the DNN model. 

\subsection{NN architecture and hyperparameters}
\label{subsec:hyperparameters}

The optimal architecture and hyperparameter for an NN of interest and selected fuel assembly, E6, are presented in Table~\ref{tab:hyperparameters}. Note that in this study, more effort was invested in optimizing the Keras DNN hyperparameter for all 32 assemblies. As indicated before, BNN VI uses the same architecture and hyperparameters as those of the Keras DNN models, while the MCD hyperparameters are obtained from our prior experience. The MATLAB ANN hyperparameters are obtained from a previous study \cite{moloko2021estimation}, briefly discussed in Subsection~\ref{subsec:matlab_ann}.      

Keras DNN hyperparameter optimization results do not display any specific pattern in terms of the hidden layers and/or hidden neurons related to their position or their surrounding environment in the core, besides that the block of hidden layers always has a pyramidal structure: the number of neurons in a layer monotonically decreases when going from the deeper layers to the shallower ones. Nevertheless, the $R^2$, so-called \emph{coefficient of determination}, obtained for the fuel assemblies, has a clearly visible tendency to decrease almost monotonically from 0.984 in the radially central region with less noisy measurements (position E6), to 0.903 in the most peripheral and, therefore, most noisy position H3. (As a reminder, $R^2$ is a measure of how well a statistical model predicts an outcome; its values are between 0 and 1, with 0 indicating that the model does not explain any variation and 1 denoting that it perfectly explains the observed variation). Values of $R^2$ for the control assemblies are in a narrower range ($0.946\pm 0.09$).  Our analysis led us to a conclusion that the hyperparameter combinations, obtained herein, do not necessarily represent the global optimum, but rather a combination that provides a satisfactory performance.


\begin{table}[htb!]
  \centering\footnotesize
  \caption{Hyperparameters for Keras DNN, MCD, BNN VI and MATLAB ANNs for fuel assembly E6.}
  \label{tab:hyperparameters}
  \begin{tabular}{lccc}
  \toprule   
   Hyperparameters         & Keras DNN           & MCD                       & MATLAB ANN\\
   \hline
   Number of inputs    & 2                  &  2                        & 1\\
   Number of outputs  & 1                  &  2\textsuperscript{*}       & 180\\
   Number of layers        & 4                  &  5                        & 1\\      
   Nodes per layer         & (157, 137, 86, 32) & (600, 500, 400, 300, 200) & 4\\
   Activation function     &  ReLU              & ReLU                      & Tanh\\   
   Learning rate           & 1.0E-04            & 1.0E-05                   &  --  \\
   Loss function           & MAE                & NLL                       & MSE\\
   Optimizer               & Adam               & Adam                      & Gradient Descent\\
                  &                &                       & Gauss-Newton\\
   Batch size              & 16                 & 180                       & --\\ 
   Number of epochs        & 750                & 1200                      & 180\\
   Dropout rate            &  --                &  0.2                      &-- \\
   Training samples        & 11047              & 11047                    & 7191\\ 
   Test samples            & 684                & 684                      & 1269\\ 
   Validation samples      & 1949               & 1949                     & --\\ 
   \bottomrule
   \textsuperscript{*}value and uncertainty      &                    &     & \\
  \end{tabular}    
\end{table}    
%


\subsection{Flux predictions with different NNs}
\label{subsec: dnn_results}

\begin{figure}[!tb]
	\centering
	\includegraphics[width=1.\linewidth]{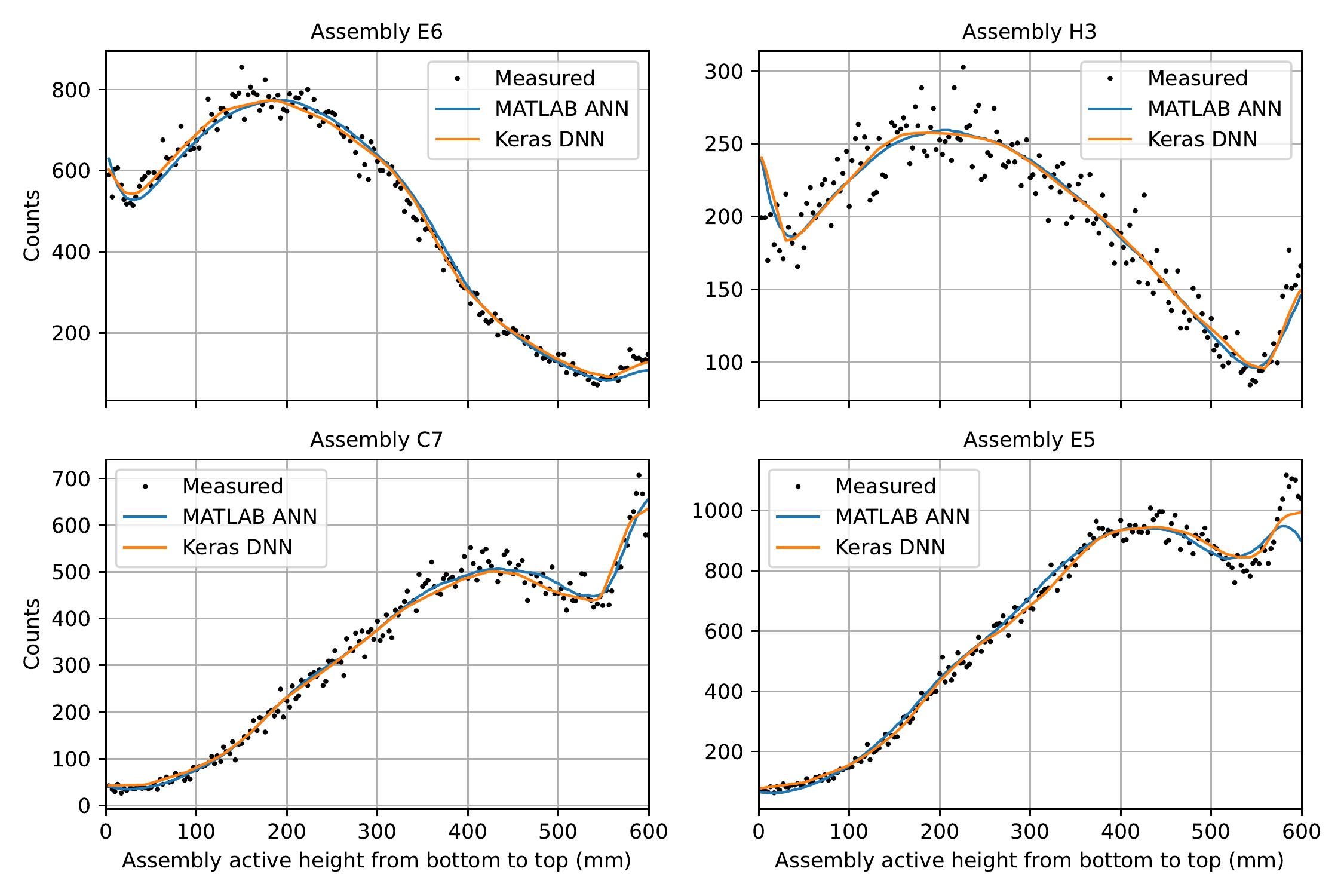}
	\caption{Measured neutron flux profiles and their predictions by MATLAB ANN and Keras DNN models, for selected fuel and control assemblies (Cycle C2105-1).}
	\label{fig:flux_profile}
\end{figure}

After training the models for all the assemblies, the 10 cycles, which were not used either in training or in the validation of the DNNs models, were utilized for flux profile predictions. Due to the large volume of generated results, only results for selected assemblies, namely C7, E5, E6 and H3 from one historical cycle (i.e.~C2105-1), are presented in this paper. A comparison of Keras DNN and MATLAB ANN flux predictions for these assemblies is shown in Figure~\ref{fig:flux_profile}. 
This comparison aims to demonstrate a number of attributes, that is, the robustness of the NN models trained on different dataset sizes, pre-processing approaches, architecture designs, and hyperparameters. 

Despite the use of different machine learning tools, it can be seen from Figure~\ref{fig:flux_profile}, that the Keras DNN and MATLAB ANN flux profile predictions are similar and that they can predict the flux profiles for unseen cycles with a good accuracy. The NNs built with both tools are also able to capture the top/bottom flux dips in the aluminium structural material of the assemblies reasonably well, except for cases where the profiles are shifted axially (these instances can be identified by the location of the top and bottom thermal peaks within the active height) and in the coupling piece, which correspond to the top-end thermal peak of the follower assemblies. 

On average, the Keras DNN slightly out-performs the MATLAB ANN (see Figures~\ref{fig:flux_profile} and \ref{fig:boxplotforassembliesmatlab}). In general, it can be concluded that the DNN model is robust, shows good prediction and generalization capability on assembly-wise comparisons for the new cycles not used in training. The predictions for all 32 assemblies, over 10 operational cycles, are presented in Figure~\ref{fig:boxplotforassembliesmatlab} in terms of the Normalized Root Mean Square Error (NRMSE), which is the Root Mean Square Error (RMSE) normalized to the mean of the measured data.

\subsection{Uncertainty quantification with MCD and BNN}
\label{subsec:UQ}

\begin{figure}[!tb]
	\centering
	\includegraphics[width=1.\linewidth]{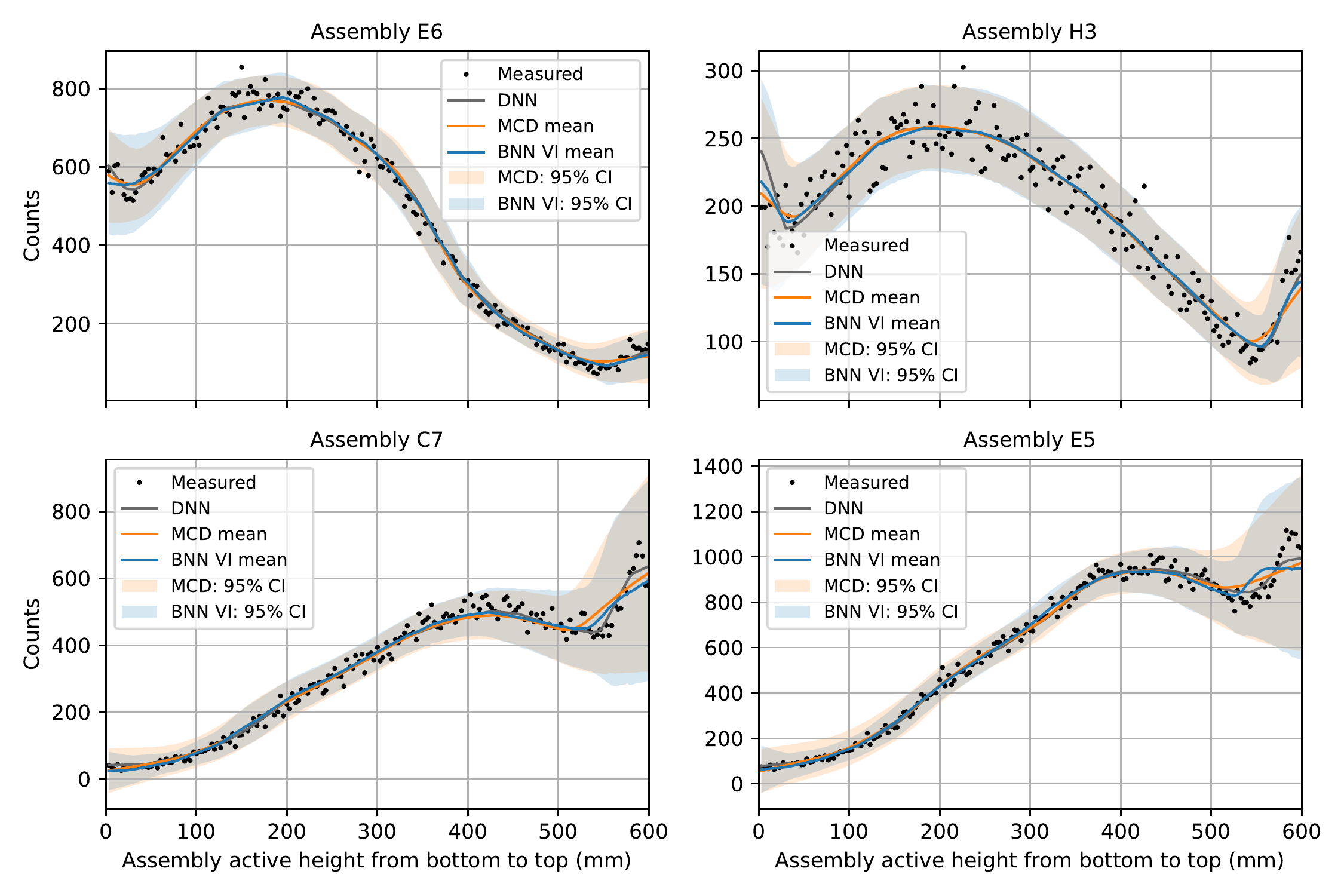}
	\caption{Measured neutron flux profiles and their predictions by different NN models along with uncertainty estimation, for selected fuel and control assemblies (Cycle C2105-1). The top two figures are for the fuel assembles and the two bottom figures are for fuel-followers in the control assemblies.}
	\label{fig: uncertainties}
\end{figure}

The design, training and most importantly, UQ in the DNN models are presented herein. The UQ in the models is performed using BNN VI and MCD models, discussed in Subsections~\ref{subsec:bnn} and \ref{subsec:mcd}. Figure~\ref{fig: uncertainties} shows measured flux profiles and NN predictions based on the DNN, BNN VI, and MCD models for two fuel (E6 and H3) and follower assemblies (C7 and E5). The UQ results, along with the mean predictions by BNN VI and MCD, are compared to the Keras DNN results that has no uncertainties. The shaded areas in these plots indicate the NN approximation uncertainties computed by BNN VI and MCD, representing the \SI{95}{\percent} Confidence Interval (CI).

\begin{figure}[!htbp]
  \centering
  \includegraphics[width=0.90\linewidth]{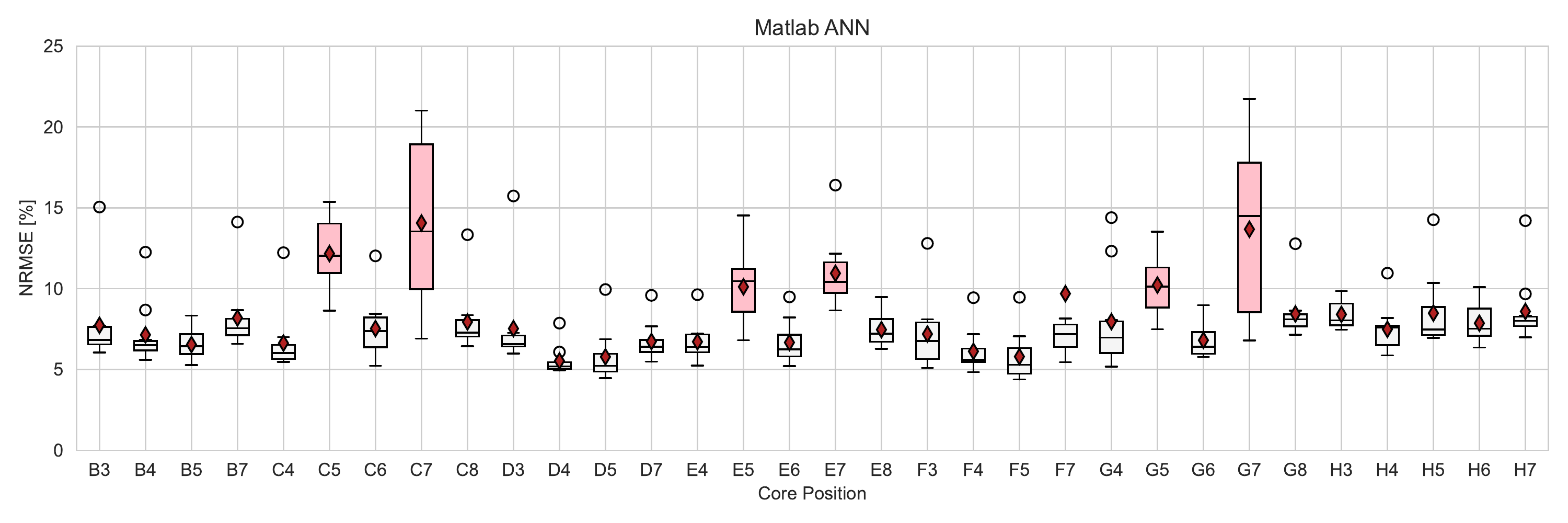}
  
  \includegraphics[width=0.90\linewidth]{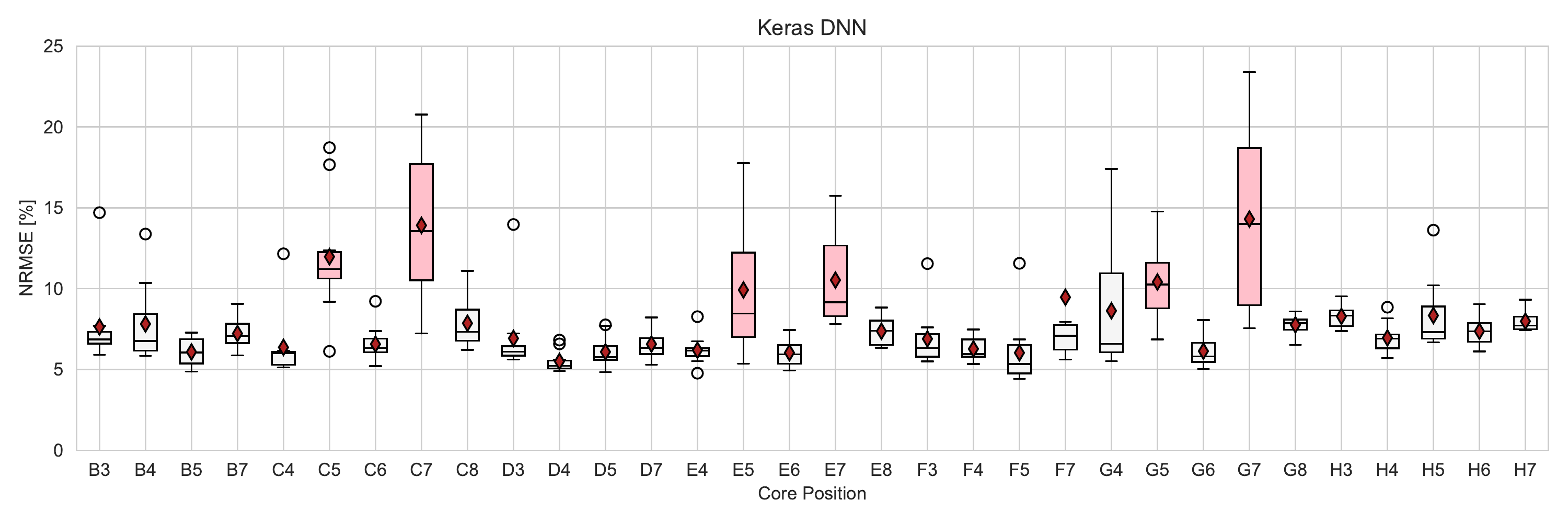}
  
  \includegraphics[width=0.90\linewidth]{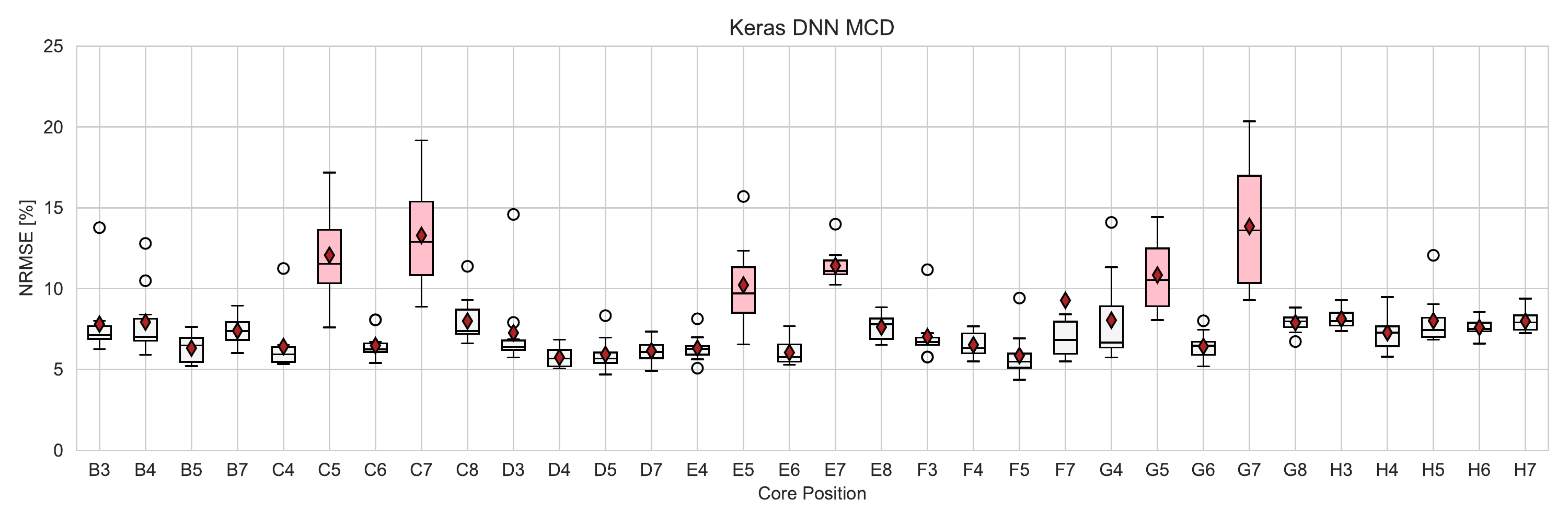}
  
  \includegraphics[width=0.90\linewidth]{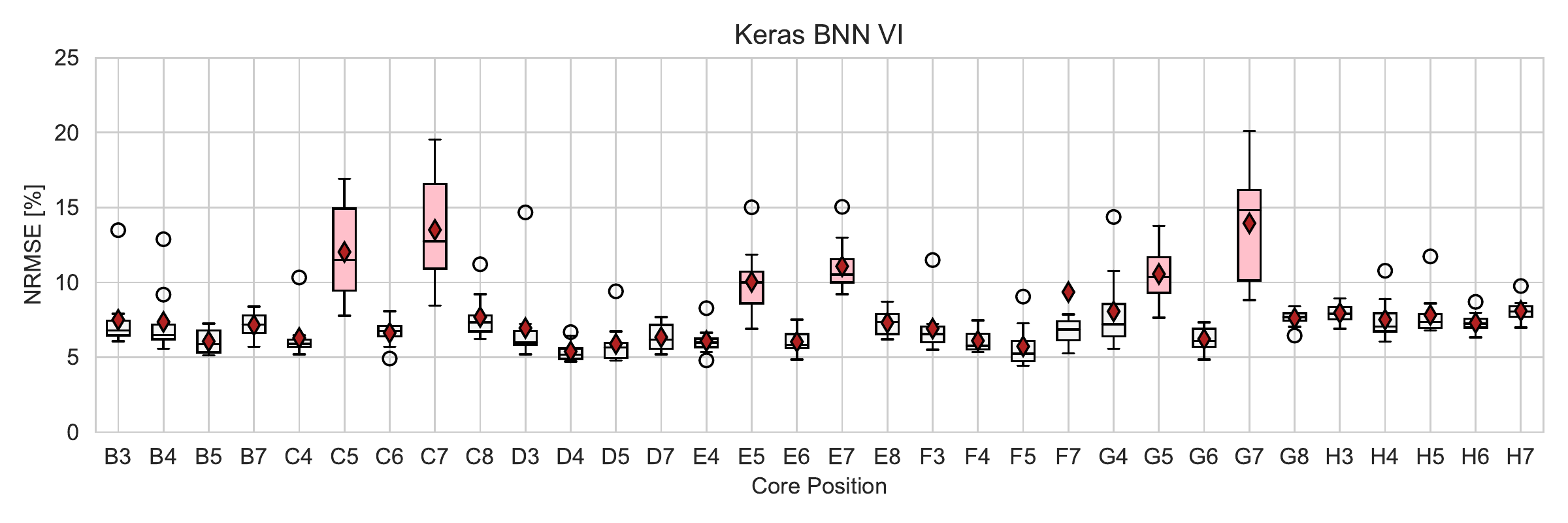}
  \caption{Distribution of the NRMSE over 10 cycles used for prediction for different assemblies and models.}%
  \label{fig:boxplotforassembliesmatlab}
\end{figure}

The Keras DNN results and the mean predictions from both BNN VI and MCD agree very well with each other, as well as to the noisy measured data of the fuel assemblies, signifying the generalization capability and robustness of the models. As was the case before, it can be seen in Figure~\ref{fig: uncertainties} for the follower assemblies, that the bottom end is not captured well in some cases while the top end thermal peak is not well resolved by the three methods. This phenomenon is observed in all follower assemblies and subsequently contributes to the large errors observed in the followers. The magnitude of the error associated with this observation shown in Figure~\ref{fig:boxplotforassembliesmatlab}, is discussed later in this paper.

The uncertainty bands from BNN VI and MCD are very similar and consistent, with the \SI{95}{\percent} CI tightly bounding all the measured points, with only a few outside the uncertainty bands. Although the mean predictions do not resolve the top end thermal peak in the follower assemblies, the uncertainty bands of BNN VI and MCD cover this data points, and thus indicate a good measure of uncertainty. The wider band of the fuel assembly H3 is attributed to its position in the core, that is, located on the periphery and thus resulting in lower counts (refer to Figure~\ref{fig:count-distribution}). The narrower and less noisy measured data signify core positions with relatively higher counts (e.g. E6). Both BNN VI and MCD are able to capture this phenomenon. 

The model prediction errors for all 32 assemblies, over 10 operational cycles (prediction dataset), are presented in Figure~\ref{fig:boxplotforassembliesmatlab} in terms of NRMSE. Both Keras DNN and MATLAB ANN models show similar trends in error distribution, with small cycle-to-cycle variations. The fuel assemblies' error lie within a range of \SI{5}{\percent} and \SI{10}{\percent}, while the follower assemblies (i.e. positions C5, C7, E5, E7, G5 and G7) show higher errors within a wider range of \SI{5}{\percent} and \SI{23}{\percent}.  The comparatively shorter and lower placed box plots, for example, assemblies C4, D4, D5, etc., indicated better prediction accuracy. 

The outliers of Figure~\ref{fig:boxplotforassembliesmatlab} were identified as defective measurements and were mainly attributed to shifted axial profiles, including the anomalies as listed in Section~\ref{subsec:preprocessing}. It can be seen that the Keras DNN has a fewer number of outliers than the MATLAB ANN. Since the performance of both models is similar, this observation is mainly attributed to training DNN with a larger and noisy dataset as opposed to the MATLAB ANN model which was trained on a smaller and smoothed dataset. The resulting Keras DNN models demonstrated some degree of robustness to these characteristics. Most interestingly, the additional effort invested in accounting for uncertainty, using BNN VI and MCD, yielded a better prediction than DNN and MATLAB ANN. The errors in both cases are reduced across all assemblies. Furthermore, BNN VI appears to be slightly more accurate than MCD.     

In summary, the DNNs and MATLAB ANN models yielded similar predictions, indicating good prediction accuracy and robustness of the models. 
Besides the excellent generalization capability, the uncertainty bands produced by MCD and BNN VI agree very well, and in general, 
they can fully envelop the noisy measurement data points.

\section{Conclusions and future work}
\label{sec:conclusions}

In this work, various neural network models have been developed to predict the assembly axial neutron flux profiles, based on the measurement data from the SAFARI-1 reactor. The training dataset consists of copper-wire activation measurements, the axial measurement locations and the measured control bank positions, obtained from the reactor’s historical cycles. A detailed study on the quantification of neural network approximation uncertainties at cycles unseen in the training process has been performed. Even though neural networks have been successfully applied to many data-driven problems in nuclear engineering, as a black-box data-driven model, they will introduce approximation uncertainties especially in generalization domains. Quantifying such approximation uncertainties is therefore important in order to improve the credibility of neural network models. In this work, Uncertainty Quantification (UQ) of the standard DNN model predictions is performed, using Monte Carlo Dropout (MCD) and Bayesian Neural Networks solved by Variational Inference (BNN VI). The analyses are conducted using Python Keras and TensorFlow Probabilistic libraries, and benchmarked by the MATLAB Neural Network Toolbox.

The standard DNNs, DNNs solved with MCD and BNNs VI results agree very well among each other as well as with the new measured dataset not used in the training process, thus indicating a good prediction and generalization capability. The DNNs and MATLAB ANN models yielded similar predictions, indicating good prediction accuracy and robustness of the models. Besides the excellent generalization capability, the uncertainty bands produced by MCD and BNN VI agree very well, and in general, they can fully envelop the noisy measurement data points. The developed ANNs are useful in supporting the experimental measurements campaign and neutronics code Verification and Validation (V\&V).


An important observation made in the current study with regards to the Keras ANNs models (i.e. DNN, MCD and BNN VI), is that MCD is much easier to implement and is able to reproduce the DNN results, including the predictive mean and model uncertainties of the BNN VI model. The MCD also reproduces the MATLAB ANN results with a good accuracy. Although the latter is user-friendly and easier to implement, its packages are not as freely available as those of Keras (Python). In the absence of MATLAB packages, it is strongly recommended to use MCD for both model predictions and UQ.

    
    

Future work includes a comparison between the neural network predictions and neutronics codes simulation. 
Also, additional research is needed to determine whether the slight differences in uncertainties produced by MCD and BNN are caused by the algorithms themselves or by the training data used in this study. Possible future studies to be considered include: (1) an attempt to use the measurement data for predicting 3D flux shapes, (2) investigate reasons for the inferior performance of NNs in the vicinity of the coupling piece of the control follower assemblies and find a way to increase the accuracy of predictions, thereby improve the overall performance of these models, and (3) benchmark with other machine learning techniques such as Gaussian Processes that can provide the interpolation uncertainties directly.


\section*{CRediT authorship contribution statement}

\textbf{L.~E.~Moloko}: Conceptualization, Methodology, Software,
Validation, Formal analysis, Investigation, Writing -- original draft,
Writing -- review \& editing, Visualization.
\textbf{P.~M.~Bokov}: Methodology, Software, Validation, Formal analysis, Investigation, Writing - original draft, Writing -- review \& editing, Visualization.
\textbf{X.~Wu}: Conceptualization, Methodology, Writing -- original draft, Writing -- review \& editing.
\textbf{K. N. Ivanov}: Conceptualization, Writing -- review \& editing.

\section*{Declaration of Competing Interest}

The authors declare that they have no known competing financial
interests or personal relationships that could have appeared
to influence the work reported in this paper.

\section*{Acknowledgement}

The authors are grateful to Ms. Hantie Labuschagne for proof reading the manuscript.


\bibliography{./bibliography.bib}

\begin{thebibliography}{10}
\expandafter\ifx\csname url\endcsname\relax
  \def\url#1{\texttt{#1}}\fi
\expandafter\ifx\csname urlprefix\endcsname\relax\def\urlprefix{URL }\fi
\expandafter\ifx\csname href\endcsname\relax
  \def\href#1#2{#2} \def\path#1{#1}\fi

\bibitem{Jaradat2016}
M.~K. Jaradat, V.~Radulovi{\'{c}}, C.~J. Park, L.~Snoj, S.~M. Alkhafaji,
  {Verification of MCNP6 model of the Jordan Research and Training Reactor
  (JRTR) for calculations of neutronic parameters}, Annals of Nuclear Energy 96
  (2016) 96--103.
\newblock \href
  {http://dx.doi.org/https://doi.org/10.1016/j.anucene.2016.06.003}
  {\path{doi:https://doi.org/10.1016/j.anucene.2016.06.003}}.

\bibitem{Snoj2011}
L.~Snoj, A.~Trkov, R.~Ja{\'{c}}imovi{\'{c}}, P.~Rogan, G.~{\v{Z}}erovnik,
  M.~Ravnik, Analysis of neutron flux distribution for the validation of
  computational methods for the optimization of research reactor utilization,
  Applied Radiation and Isotopes 69~(1) (2011) 136--141.
\newblock \href
  {http://dx.doi.org/https://doi.org/10.1016/j.apradiso.2010.08.019}
  {\path{doi:https://doi.org/10.1016/j.apradiso.2010.08.019}}.

\bibitem{Haffner2019}
R.~Haffner, W.~H. Miller, S.~Morris,
  \href{https://doi.org/10.1016/j.apradiso.2019.05.001}{{Verification of I-131
  yields from the neutron irradiation of tellurium}}, Applied Radiation and
  Isotopes 151~(May) (2019) 52--61.
\newline\urlprefix\url{https://doi.org/10.1016/j.apradiso.2019.05.001}

\bibitem{international2019iaea}
\href{https://www.iaea.org/publications/13547/benchmarking-against-experimental-data-of-neutronics-and-thermohydraulic-computational-methods-and-tools-for-operation-and-safety-analysis-of-research-reactors}{Benchmarking
  Against Experimental Data of Neutronics and Thermohydraulic Computational
  Methods and Tools for Operation and Safety Analysis of Research Reactors},
  no. 1879 in TECDOC Series, INTERNATIONAL ATOMIC ENERGY AGENCY, 2019.
\newline\urlprefix\url{https://www.iaea.org/publications/13547/benchmarking-against-experimental-data-of-neutronics-and-thermohydraulic-computational-methods-and-tools-for-operation-and-safety-analysis-of-research-reactors}

\bibitem{Dias2016a}
A.~M. Dias, F.~C. Silva,
  \href{http://dx.doi.org/10.1016/j.anucene.2015.12.002}{Determination of the
  power density distribution in a {PWR} reactor based on neutron flux
  measurements at fixed reactor incore detectors}, Annals of Nuclear Energy 90
  (2016) 148--156.
\newline\urlprefix\url{http://dx.doi.org/10.1016/j.anucene.2015.12.002}

\bibitem{moloko2016benchmark}
L.~E. Moloko, J.~Politello, J.~Di~Salvo, C.~D'Aletto, The {2D} static benchmark
  calculations for the {SAFARI-1} research reactor core characterisation, in:
  Physics of Reactors 2016: Unifying Theory and Experiments in the 21st
  Century, PHYSOR 2016, American Nuclear Society, Sun Valley Resort, Idaho,
  USA, 2016, pp. 2069--2081.

\bibitem{international2015iaea}
\href{https://www.iaea.org/publications/10578/research-reactor-benchmarking-database-facility-specification-and-experimental-data}{Research
  Reactor Benchmarking Database: Facility Specification and Experimental Data},
  no. 480 in Technical Reports Series, INTERNATIONAL ATOMIC ENERGY AGENCY,
  Vienna, 2015.
\newline\urlprefix\url{https://www.iaea.org/publications/10578/research-reactor-benchmarking-database-facility-specification-and-experimental-data}

\bibitem{Demuth2014}
H.~B. Demuth, M.~H. Beale, O.~De~Jess, M.~T. Hagan, Neural Network Design, 2nd
  Edition, Martin Hagan, Stillwater, OK, USA, 2014.

\bibitem{Suzuki2011}
K.~Suzuki, Artificial Neural Networks -- Industrial and Control Engineering
  Applications, IntechOpen, Rijeka, 2011.

\bibitem{Souza2006a}
R.~Souza, J.~Moreira, Power peak factor for protection systems -- experimental
  data for developing a correlation, Annals of Nuclear Energy 33 (2006)
  609--621.

\bibitem{Souza2006b}
R.~M. G.~P. Souza, J.~M.~L. Moreira, Neural network correlation for power peak
  factor estimation, Annals of Nuclear Energy 33~(7) (2006) 594--608.

\bibitem{Terman2018}
M.~S. Terman, N.~M. Kojouri, H.~Khalafi, Determination of control rod positions
  during fuel life-cycle using fixed in-core self-powered neutron detectors of
  {Tehran Research Reactor}, Nuclear Engineering and Design 331 (2018) 68--82.

\bibitem{Schlunz2015c}
E.~B. Schl\"{u}nz, P.~M. Bokov, J.~H. Van~Vuuren, Application of artificial
  neural networks for predicting core parameters for the {SAFARI-1} nuclear
  research reactor, in: Proceedings of the 44th Annual Conference of the
  Operations Research Society of South Africa {(ORSSA 2015)}, Hartbeespoort,
  South Africa, 13--16 September, 2015, pp. 12--22.

\bibitem{Schlunz2016b}
E.~B. Schl\"{u}nz, P.~M. Bokov, J.~H. Van~Vuuren, A comparative study on
  multiobjective metaheuristics for solving constrained in-core fuel management
  optimisation problems, Computers \& Operations Research 75 (2016) 174--190.

\bibitem{Schlunz2018}
E.~B. Schl\"{u}nz, P.~M. Bokov, J.~H. Van~Vuuren, Multiobjective in-core
  nuclear fuel management optimisation by means of a hyperheuristic, Swarm and
  Evolutionary Computation 42 (2018) 58--76.

\bibitem{Rabie2020}
A.~S. Rabie, M.~I. Radaideh, T.~Kozlowski,
  \href{https://www.sciencedirect.com/science/Article/pii/S1738573320301637}{Application
  of deep neural networks for high-dimensional large {BWR} core neutronics},
  Nuclear Engineering and Technology 52~(12) (2020) 2709--2716.
\newblock \href {http://dx.doi.org/https://doi.org/10.1016/j.net.2020.05.010}
  {\path{doi:https://doi.org/10.1016/j.net.2020.05.010}}.
\newline\urlprefix\url{https://www.sciencedirect.com/science/Article/pii/S1738573320301637}

\bibitem{koo2019}
Y.~D. Koo, Y.~J. An, C.-H. Kim, M.~G. Na, Nuclear reactor vessel water level
  prediction during severe accidents using deep neural networks, Nuclear
  Engineering and Technology 51~(3) (2019) 723--730.
\newblock \href {http://dx.doi.org/https://doi.org/10.1016/j.net.2018.12.019}
  {\path{doi:https://doi.org/10.1016/j.net.2018.12.019}}.

\bibitem{gal2016dropout}
Y.~Gal, Z.~Ghahramani, Dropout as a {Bayesian} approximation: Representing
  model uncertainty in {Deep Learning}, in: International Conference on Machine
  Learning, PMLR, 2016, pp. 1050--1059.

\bibitem{blundell2015weight}
C.~Blundell, J.~Cornebise, K.~Kavukcuoglu, D.~Wierstra, Weight uncertainty in
  neural network, in: International Conference on Machine Learning, PMLR, 2015,
  pp. 1613--1622.

\bibitem{Srivastava2014dropout}
N.~Srivastava, G.~Hinton, A.~Krizhevsky, I.~Sutskever, R.~Salakhutdinov,
  Dropout: A simple way to prevent neural networks from overfitting, Journal of
  Machine Learning Research 15~(1) (2014) 1929--1958.

\bibitem{moloko2021estimation}
L.~E. Moloko, P.~M. Bokov, K.~N. Ivanov, Estimation of the axial neutron flux
  profiles in the {SAFARI-1} core using artificial neural networks, in:
  Proceedings of the 2021 International Conference on Mathematics and
  Computational Methods Applied to Nuclear Science and Engineering (M\&C-2021),
  Raleigh, North Carolina, October 3--7, 2021, 2021, pp. 1644--1653.

\bibitem{moloko2022UQ}
L.~E. Moloko, P.~M. Bokov, X.~Wu, K.~N. Ivanov, Quantification of neural
  networks uncertainties with applications to {SAFARI-1} axial neutron flux
  profiles, in: Proceedings of the International Conference on Physics of
  Reactors (PHYSOR 2022), ANS, 2022, pp. 1398--1407.

\bibitem{international2022iaea}
\href{https://www.iaea.org/publications/15041/benchmarks-of-fuel-burnup-and-material-activation-computational-tools-against-experimental-data-for-research-reactors}{Benchmarks
  of Fuel Burnup and Material Activation Computational Tools Against
  Experimental Data for Research Reactors}, no. 1992 in TECDOC Series,
  INTERNATIONAL ATOMIC ENERGY AGENCY, 2022.
\newline\urlprefix\url{https://www.iaea.org/publications/15041/benchmarks-of-fuel-burnup-and-material-activation-computational-tools-against-experimental-data-for-research-reactors}

\bibitem{Garis1998}
N.~S. Garis, I.~Pázsit, U.~Sandberg, T.~Andersson, Determination of {PWR}
  control rod position by core physics and neural network methods, Nuclear
  Technology 123~(3) (1998) 278--295.
\newblock \href {http://dx.doi.org/10.13182/NT98-A2899}
  {\path{doi:10.13182/NT98-A2899}}.

\bibitem{Savitzky1964}
A.~Savitzky, M.~J.~E. Golay, Smoothing and differentiation of data by
  simplified least squares procedures., Analytical Chemistry 36~(8) (1964)
  1627--1639.
\newblock \href {http://dx.doi.org/10.1021/ac60214a047}
  {\path{doi:10.1021/ac60214a047}}.

\bibitem{Schafer2011}
R.~Schafer, What is a {Savitzky-Golay} filter? [lecture notes], IEEE Signal
  Processing Magazine 28 (2011) 111--117.
\newblock \href {http://dx.doi.org/10.1109/MSP.2011.941097}
  {\path{doi:10.1109/MSP.2011.941097}}.

\bibitem{Jain2005}
A.~Jain, K.~Nandakumar, A.~Ross, Score normalization in multimodal biometric
  system, Pattern Recognition 38 (2005) 2270--2285.

\bibitem{Nawi2013}
N.~M. Nawi, W.~H. Atomi, M.~Z. Rehman, The effect of data pre-processing on
  optimized training of artificial neural networks, Procedia Technology 11
  (2013) 32--39, 4th International Conference on Electrical Engineering and
  Informatics, ICEEI 2013.

\bibitem{Yu2006}
L.~Yu, S.~Wang, K.~K. Lai, An integrated data preparation scheme for neural
  network data analysis., Knowledge and Data Engineering, IEEE Transactions on
  18 (2006) 217--230.
\newblock \href {http://dx.doi.org/10.1109/TKDE.2006.22}
  {\path{doi:10.1109/TKDE.2006.22}}.

\bibitem{Cleveland1979}
W.~S. Cleveland, Robust locally weighted regression and smoothing scatterplots,
  Journal of the American Statistical Association 74~(368) (1979) 829--836.
\newblock \href {http://dx.doi.org/10.1080/01621459.1979.10481038}
  {\path{doi:10.1080/01621459.1979.10481038}}.

\bibitem{PriceD2022}
D.~Price, M.~I. Radaideh, B.~Kochunas, Multiobjective optimization of nuclear
  microreactor reactivity control system operation with swarm and evolutionary
  algorithms, Nuclear Engineering and Design 393 (2022) 111776.
\newblock \href
  {http://dx.doi.org/https://doi.org/10.1016/j.nucengdes.2022.111776}
  {\path{doi:https://doi.org/10.1016/j.nucengdes.2022.111776}}.

\bibitem{Hornik1989}
K.~Hornik, M.~Stinchcombe, H.~White, Multilayer feedforward networks are
  universal approximators, Neural Networks 2~(5) (1989) 359--366.

\bibitem{Lecun2015Deeplearning}
Y.~LeCun, Y.~Bengio, G.~Hinton, Deep learning, Nature 521 (2015) 436--44.

\bibitem{Saptarshi2020Deeplearning}
S.~Sengupta, S.~Basak, P.~Saikia, S.~Paul, V.~Tsalavoutis, F.~Atiah, V.~Ravi,
  A.~Peters, A review of deep learning with special emphasis on architectures,
  applications and recent trends, Knowledge-Based Systems 194 (2020) 105596.

\bibitem{Gal2016Uncertainty}
Y.~Gal, Uncertainty in deep learning, Ph.D. thesis, University of Cambridge
  (2016).

\bibitem{bachstein2019uncertainty}
S.~Bachstein, Uncertainty quantification in deep learning, Master Thesis.

\bibitem{Liu2018DNN}
Y.~Liu, N.~Dinh, Y.~Sato, B.~Niceno, Data-driven modeling for boiling heat
  transfer: Using deep neural networks and high-fidelity simulation results,
  Applied Thermal Engineering 144 (2018) 305--320.
\newblock \href
  {http://dx.doi.org/https://doi.org/10.1016/j.applthermaleng.2018.08.041}
  {\path{doi:https://doi.org/10.1016/j.applthermaleng.2018.08.041}}.

\bibitem{goan2020bayesian}
E.~Goan, C.~Fookes, Bayesian neural networks: An introduction and survey, in:
  Case Studies in Applied Bayesian Data Science, Springer, Cham, 2020, pp.
  45--87.

\bibitem{neal2012bayesian}
R.~M. Neal, Bayesian learning for neural networks, Vol. 118, Springer Science
  \& Business Media, 2012.

\bibitem{blei2017variational}
D.~M. Blei, A.~Kucukelbir, J.~D. McAuliffe, Variational inference: A review for
  statisticians, Journal of the American Statistical Association 112~(518)
  (2017) 859--877.

\bibitem{tzikas2008variational}
D.~G. Tzikas, A.~C. Likas, N.~P. Galatsanos, The variational approximation for
  bayesian inference, IEEE Signal Processing Magazine 25~(6) (2008) 131--146.

\bibitem{chollet2015keras}
F.~Chollet, \href{https://github.com/fchollet/keras}{Keras} (2015).
\newline\urlprefix\url{https://github.com/fchollet/keras}

\bibitem{neorl2021}
M.~Radaideh, K.~Du, P.~Seurin, D.~Seyler, X.~Gu, H.~Wang, K.~Shirvan,
  \href{https://arxiv.org/abs/2112.07057}{Neorl: Neuroevolution optimization
  with reinforcement learning} (2021).
\newblock \href {http://dx.doi.org/10.48550/ARXIV.2112.07057}
  {\path{doi:10.48550/ARXIV.2112.07057}}.
\newline\urlprefix\url{https://arxiv.org/abs/2112.07057}

\bibitem{BergstraB2012}
J.~Bergstra, Y.~Bengio, Random search for hyper-parameter optimization.,
  Journal of Machine Learning Research 13 (2012) 281--305.

\bibitem{MATLAB2017a}
The Mathworks, Inc., Natick, Massachusetts, {MATLAB Version 9.2.0.959691
  (R2017a) Update 3} (2017).

\end{thebibliography}

\end{document}